\pdfoutput=1

\documentclass[11pt]{article}

\usepackage{EMNLP2022}

\usepackage{times}
\usepackage{latexsym}

\usepackage[T1]{fontenc}

\usepackage[utf8]{inputenc}

\usepackage{microtype}

\usepackage{inconsolata}

\usepackage{microtype}
\usepackage{amssymb}
\usepackage{graphicx}
\usepackage{multirow}
\usepackage{caption}
\usepackage{amsmath}
\usepackage{rotating}

\usepackage{verbatim}
\usepackage{xcolor,pifont}
\newcommand*\colourcheck[1]{%
  \expandafter\newcommand\csname #1check\endcsname{\textcolor{#1}{\ding{52}}}%
}
\colourcheck{green}
\newcommand{\xmark}{\textcolor{red}{\ding{55}}}
\title{That's the Wrong Lung! Evaluating and Improving the Interpretability \\ of Unsupervised Multimodal Encoders for Medical Data}

\author{Denis Jered McInerney \\
  Northeastern University \\
  \texttt{mcinerney.de@northeastern.edu} \\\And
  Geoffrey Young \\
  Brigham and Women's Hospital \\
  \texttt{gsyoung@bwh.harvard.edu} \\\AND
  Jan-Willem van de Meent \\
  University of Amsterdam \\
  \texttt{j.w.vandemeent@uva.nl} \\\And
  Byron C. Wallace \\
  Northeastern University \\
  \texttt{b.wallace@northeastern.edu} \\}

\begin{document}
\maketitle
\begin{abstract}
    Pretraining multimodal models on Electronic Health Records (EHRs) provides a means of learning representations 
    that can transfer to downstream tasks with minimal supervision.
    Recent multimodal models 
    induce soft local alignments between image regions and sentences. 
    This is of particular interest in the medical domain, where alignments might highlight regions in an image relevant to specific phenomena described in free-text. 
    While past work has suggested that attention ``heatmaps'' can be interpreted in this manner, 
    there has been little 
    evaluation of such alignments. 
    We compare alignments from a state-of-the-art multimodal (image and text) model for EHR with human annotations that link image regions to sentences. 
    Our main finding is that the text has an often weak or unintuitive influence on attention; 
    alignments do not consistently reflect basic anatomical information. Moreover, synthetic modifications --- such as substituting ``left'' for ``right'' --- do not substantially influence highlights.
    Simple techniques such as allowing the model to opt out of attending to the image and few-shot finetuning 
    show promise in terms of their ability to improve 
    alignments with very little or no supervision.
    We make our code and checkpoints open-source.\footnote{\url{https://github.com/dmcinerney/gloria}}
\end{abstract}

\section{Introduction}

There has been a flurry of recent work on model architectures and self-supervised training objectives for multimodal representation learning, both generally \cite{Li2019VisualBERTAS, Tan2019LXMERTLC, Huang2020PixelBERTAI, Su2020VLBERTPO, Chen2020UNITERUI} and for medical data specifically \cite{Wang2018TieNetTE, Chauhan2020JointMO, Li2020ACO}.
These methods 
yield representations that permit efficient learning on various multimodal downstream tasks (e.g., classification, captioning).


\begin{figure*}
    \centering
    \includegraphics[scale=.5]{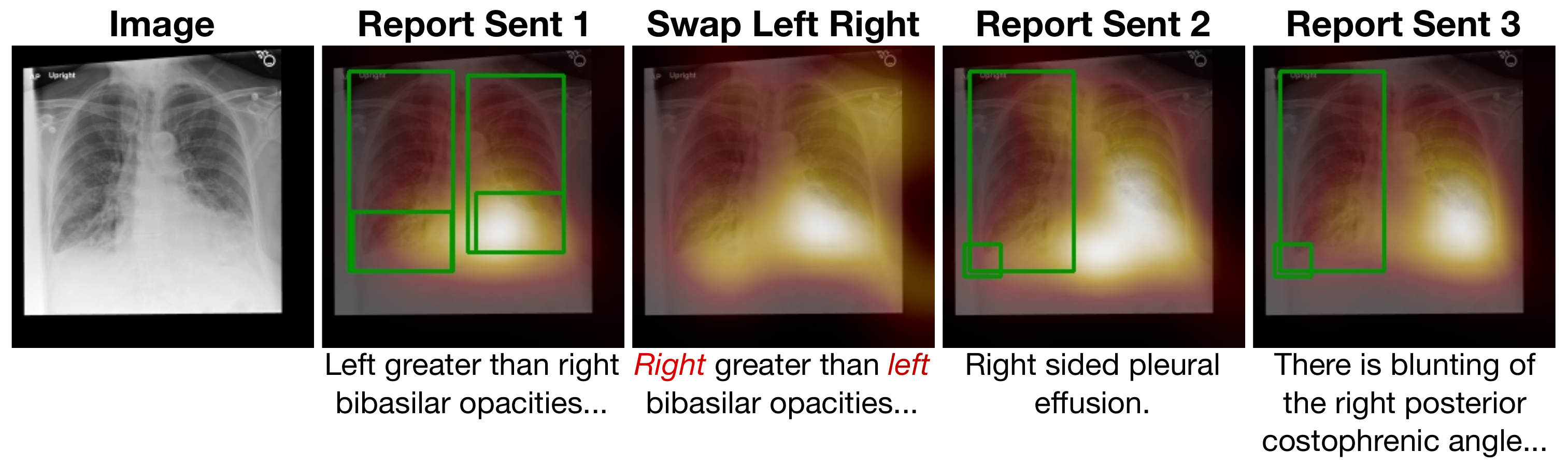}
    \caption{Alignment failures often occur when the model (overly) focuses on one aspect of the image, largely ignoring the text. (Note: images are ``mirrored'', so right and left are flipped.)}
    \label{fig:subtle_failure_case}
\end{figure*}

Given the inherently multimodal nature of much medical data --- e.g., in radiology images and text are naturally paired --- there has been particular interest in designing multimodal models for Electronic Health Records (EHRs) data.
However, one of the factors that currently stands in the way of broader adoption is interpretability.
Neural models that map image-text pairs to shared representations are opaque. 
Consequently, doctors have no way of knowing whether such models rely on meaningful clinical signals or data artifacts \cite{10.1371/journal.pmed.1002683}. 

Recent work has proposed models that soft-align text snippets to image regions.
This may afford a type of interpretability by allowing practitioners to inspect what the model has ``learned'' or allow more efficient identification of relevant regions.
Past work has presented 
illustrative multimodal ``saliency'' maps in which such models highlight plausible regions. 
But such highlights also risk providing a false sense that the model ``understands'' more than it actually does, and irrelevant highlights would be antithetical to the goal of a efficiency in clinical decision support.

Multimodal models may fail in a few obvious ways; they may focus on the wrong part of an image, fail to localize by producing a high-entropy attention distribution, or localize too much and miss a larger region of interest. 
However, even when image attention \emph{appears} reasonable, it may not in actuality reflect both modalities.
Figure \ref{fig:subtle_failure_case} shows an example. 
Here 
the model ostensibly succeeds at identifying the image region relevant to the given text (left). 
One may be tempted to conclude the model has ``understood'' the text and indicated the corresponding region.
But this may be misleading: We can see that the same model yields a similar attention pattern when provided text with radically different semantics (e.g., when swapping ``right'' with ``left''), or when providing sentences referencing an abnormality in another region.
Our contributions are as follows.
(i) We appraise the interpretability of soft-alignments induced between images and texts by existing neural multimodal models for radiology, both retrospectively and via manual radiologist assessments.
To the best of our knowledge, this is the first such evaluation.
(ii) We propose methods that improve the ability of multimodal models for EHR to intuitively align image regions with texts. 

\section{Preliminaries}

We aim to evaluate the localization abilities of multimodal models for EHR. 
For this we focus on the recently proposed GLoRIA model \cite{Huang_2021_ICCV}, which is representative of state-of-the-art, 
transformer-based multimodal architectures and accompanying pre-training methods.
For completeness we also analyze (a modified version of) UNITER \cite{Chen2020UNITERUI}.
We next review details of these models, and then discuss the datasets we use to evaluate the alignments they induce.

\subsection{GLoRIA}

GLoRIA uses Clinical BERT \cite{Alsentzer2019PubliclyAC} as a text encoder 
and ResNet \cite{He2016DeepRL} as 
an image encoder.
Unlike prior work, GLoRIA does not assume an image can be partitioned into different objects, which is important because pre-trained object detectors are not readily available for X-ray images. 
GLoRIA passes a CNN over the 
image to yield local region representations. 
This is useful because a finding within an X-ray described in a report will usually appear in only a small region of the corresponding image \cite{Huang_2021_ICCV}.
GLoRIA 
exploits this intuition via a local contrastive loss term in the objective.

We assume a dataset of instances comprising an image $x_v$ and a sentence from the corresponding report $x_t$, 
and the model consumes this to produce a set of local embeddings and a global embedding per modality: $v_l \in \mathbb{R}^{M \times D}$, $v_g \in \mathbb{R}^D$, $t_l \in \mathbb{R}^{N \times D}$, and $t_g \in \mathbb{R}^D$.
To construct the local contrastive loss, an attention mechanism \cite{bahdanau2014neural} is applied to local 
image embeddings, queried by the local text embeddings. This 
induces a soft alignment between the local vectors of each mode:
\begin{equation}
    a_{ij} = \frac{\exp{(t_{li}^T v_{lj} / \tau)}}{\sum_{k=1} ^ M \exp{(t_{li}^T v_{lk} / \tau)}}
\end{equation}
where $t_i$ is the $i$th text embedding, $v_j$ the $j$th image embedding, and $\tau$ is a temperature hyperparameter.

\subsection{UNITER}\label{sec:uniter}

Despite the challenges inherent to adopting "general-domain" multimodal models for this domain (discussed in Appendix \ref{sec:general_domain_challenges}), 
we 
modify UNITER to serve as an additional 
model for analysis. 
We provide details regarding how we have implemented UNITER in Appendix \ref{sec:uniter-details}, but note here that \emph{this requires ground-truth bounding boxes as inputs}, which means that (a) results with respect to most metrics (which measure overlap with target bounding boxes) for UNITER will be artificially high, and, (b) we could not use this method in practice, because it requires a set of reference bounding boxes as input (including at inference time). 
We include this for completeness.

\subsection{Data and Metrics}\label{sec:data_and_metrics}


\vspace{0.35em}
\noindent\textbf{Data}
Our retrospective evaluation of localization abilities is made possible by the MIMIC-CXR \cite{Johnson2019MIMICCXRAD, Johnson2019MIMICCXRAL} and Chest ImaGenome \cite{Wu2021ChestID} datasets.
MIMIC-CXR comprises chest X-rays and corresponding radiology reports.
ImaGenome 
includes 1000 
manually annotated image/report pairs,\footnote{Annotations were automatically derived then cleaned.} 
with bounding boxes for anatomical locations, 
links between referring sentences and image bounding boxes, and a set of conditions and positive/negative context annotations\footnote{Here, context refers to whether the condition is negated in the text (negative) or not (positive).} associated with each sentence/bounding box pair.


\vspace{0.35em}
\noindent\textbf{Metrics}
We quantify 
the degree to which attention 
highlights the region to which a text snippet refers by comparing average attention over an input sentence $x_j = \frac{1}{N}\sum_{i=1}^N a_{ij}$ with reference annotated bounding boxes associated with the sentence.

We use several metrics to measure the alignment between soft attention weights and bounding boxes. 
We create scores $s \in \mathbb{R}^P$ for each of the $P$ pixels based on the attention weight assigned to the image region the pixel belongs to.
Specifically, for GLoRIA we use upsampling with bilinear interpolation to distribute attention over pixels.
For UNITER, we score pixels by taking a max over attention scores for the bounding boxes that contain the pixel (scores for pixels not in any bounding boxes are 0).
We use bounding boxes to create a segmentation label $\ell \in \mathbb{R}^P$ where $\ell_p=1$ if pixel $i$ is in any of the bounding boxes, and $\ell_p=0$ otherwise.
Given pixel-level scores $s$ and pixel-level segmentation labels $\ell_p$, we can compute the \textbf{AUROC}, \textbf{Average Precision}, and \textbf{Intersection Over Union} (IOU) at varying pixel percentile thresholds for the ranking ordered by $s$ (See section \ref{sec:metric_demo}).

We 
also adopt a simple, interpretable metric to capture the accuracy of similarity scores assigned to pairs of images and texts.
Specifically, we use a simpler version of the text retrieval task from \cite{Huang_2021_ICCV}: 
We report the percentage of time the similarity between an image and a sentence from the corresponding report is greater than the similarity between the image and a random sentence taken from a different report in the dataset. 
This allows us to interpret 50\% as the mean value of a totally random similarity measure.

\section{Are Alignment Weights Accurate?}\label{sec:alignment_accuracy}


\begin{table}[]
    \centering
    \begin{tabular}{c c c c}
        AUROC & Avg. P & IOU@5/10/30\% \\
        \hline
69.07 & 51.68 & 3.79/6.69/20.10 \\
    \end{tabular}
    \caption{Localization performance of GLoRIA.} 
    \label{tab:gloria_localization_performance}
\end{table}
\vspace{-5pt}

We first use the metrics defined above to 
evaluate the pretrained, publicly available weights for GLoRIA \cite{Huang_2021_ICCV}.
Table \ref{tab:gloria_localization_performance} reports the metrics used to evaluate localization on the gold split of the ImaGenome dataset.

AUROC scores are well over 50\%, suggesting reasonable localization performance.
IOU scores are 
small, which is expected 
as target bounding boxes tend to be much larger than the actual regions of interest and serve more to detect errors when highlighted regions are far from where they should be; this is further supported by the relatively high average precision scores.\footnote{In Section \ref{sec:dropping_large_bboxes}, we address this with a modified evaluation that drops some large bounding boxes in the labels.}
However, while seemingly promising, our results below suggest that the attention patterns here may be less multimodal than one might expect.

\begin{figure}
    \centering
    \includegraphics[scale=.66]{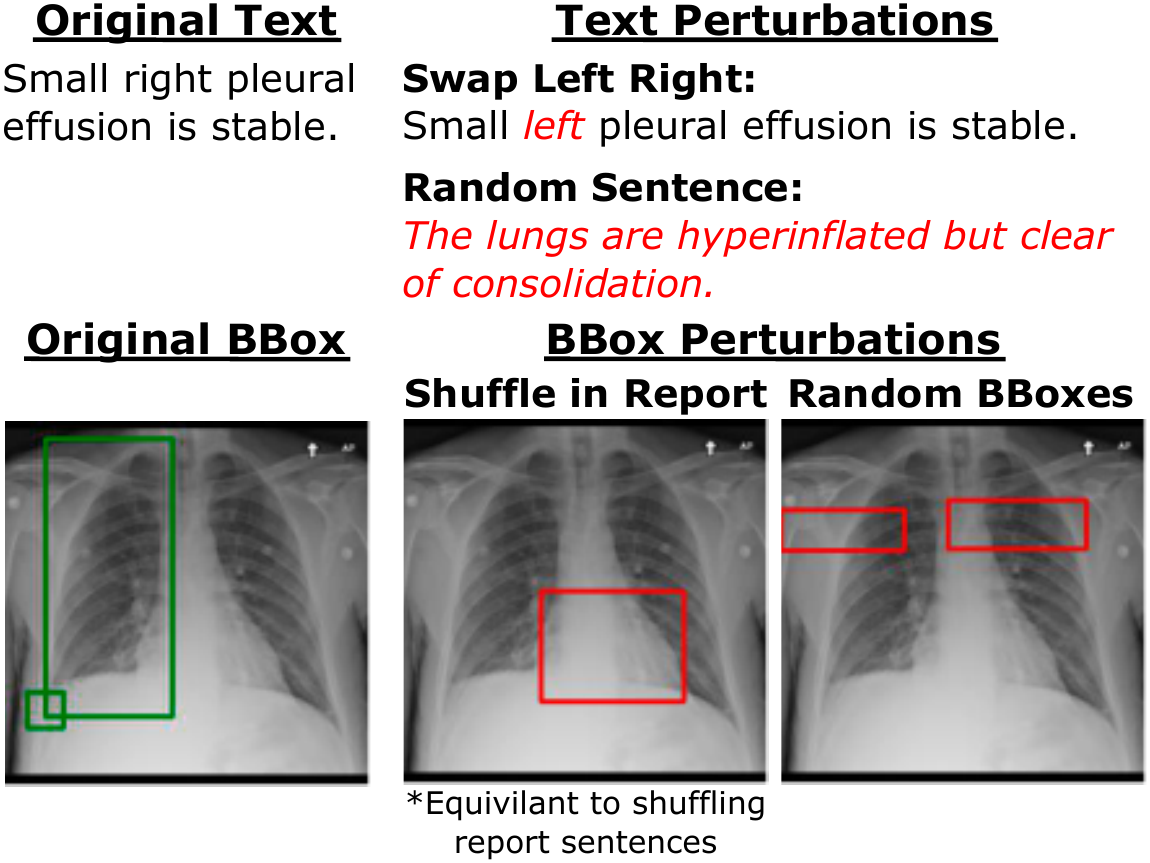}
    \caption{Examples of each \textbf{perturbation} for a given instance. (Synth w/ Swapped Conditions example in Appendix \ref{fig:perturbations_app}.)}
    \label{fig:perturbations}
\end{figure}

We 
next focus on evaluating the degree to which these patterns actually reflect the associated text.
To this end we 
perturb 
instances in ways that ought to shift the attention pattern (Section \ref{sec:perturbations}), e.g., by replacing ``right'' with ``left'' in the text.
We then 
identify data subsets in Section \ref{sec:subsets} 
comprising ``complex'' instances, where we expect the image and text to be closely correlated at a local level.


\subsection{Perturbations}\label{sec:perturbations}

Figure \ref{fig:perturbations} shows examples of the perturbations that include: Swapping ``left'' with ``right'' ({\bf Swap Left Right}); Shuffling the target bounding boxes for sentences within the same report at random ({\bf Shuffle in Report}); Replacing \emph{sentences} in a report with other sentences, randomly drawn from the rest of the dataset ({\bf Random Sentences}); Replacing target \emph{bounding boxes} with other bounding boxes randomly sampled from the dataset ({\bf Random BBoxes})\footnote{\textbf{Shuffle in Report} bboxes will still correspond to valid and noteworthy anatomical regions, but \textbf{Random BBoxes} bboxes will not correspond to valid anatomical regions at all.}, and; Swapping the correct conditions in a synthetically created prompt with random conditions \textbf{Synth w/ Swapped Conditions}. We include additional details about synthetic sentences and perturbations in Appendices~\ref{sec:synthetic_sents} and~\ref{sec:perturbations-details}.

Under these perturbations, we would expect a well-behaved model to shift its attention distribution over the image accordingly, resulting in a decrease in localization scores (overlap with the original reference bounding boxes).
The {\bf Random BBoxes} perturbation in particular targets the degree to which the attention relies specifically on the image modality, because here the ``target'' bounding boxes have been replaced with bounding boxes associated with \emph{random other images}.
By contrast, all other perturbations should measure the degree to which the model is sensitive to changes to the text (even {\bf Shuffle in Report}, which is equivalent to shuffling the sentences in a report).

If attention maps reflect alignments with input texts, 
then under these perturbations one should expect large negative differences in performance ($\Delta$metric) relative to observed performance using the unperturbed data.
For all but {\bf Random BBoxes}, if the performance does not much change ($\Delta$metric $\approx$ 0), this suggests the attention maps are somewhat invariant to the text modality.


\subsection{Subsets}\label{sec:subsets}

We perform granular evaluations using specific data subsets, including:
(1) \textbf{Abnormal} instances with an abnormality, (2) \textbf{One Lung} instances with only one side of the Chest X-ray (left or right) referenced, and (3) \textbf{Most Diverse Report BBoxes (MDRB)} instances with a lot of diversity in the labels for sentences in the same report.
Details are in Appendix \ref{sec:subset_details}.

Intuitively, some of the perturbations in Section \ref{sec:perturbations} should mainly effect certain subsets: \textbf{Swap Left Right} should most impact the \textbf{One Lung} subset, \textbf{Shuffle in Report} should mainly effect \textbf{MDRB}, and \textbf{Random Sentences}, \textbf{Random BBoxes}, and \textbf{Synth w/ Swapped Conditions} should primarily effect \textbf{Abnormal} examples.


\subsection{Annotations for Post-hoc Evaluation}

We enlist a domain expert (radiologist) to conduct annotations to complement our retrospective quantitative evaluations. 
We elicit judgements on a five-point Likert scale regarding the recall, precision, and ``intuitiveness'' of image highlights induced for text snippets.\footnote{For recall and precision, points on the Likert scale are intended to correspond to buckets of 0-20, 20-40, 40-60, 60-80, and 80-100 percent respectively.}   
More details are in the Appendix, including 
annotation instructions (Section \ref{sec:annotations_appendix}) and a 
screenshot of the interface (Figure \ref{fig:interface}).  

\subsection{Results}

\begin{figure}
    \centering
    \includegraphics[scale=.75]{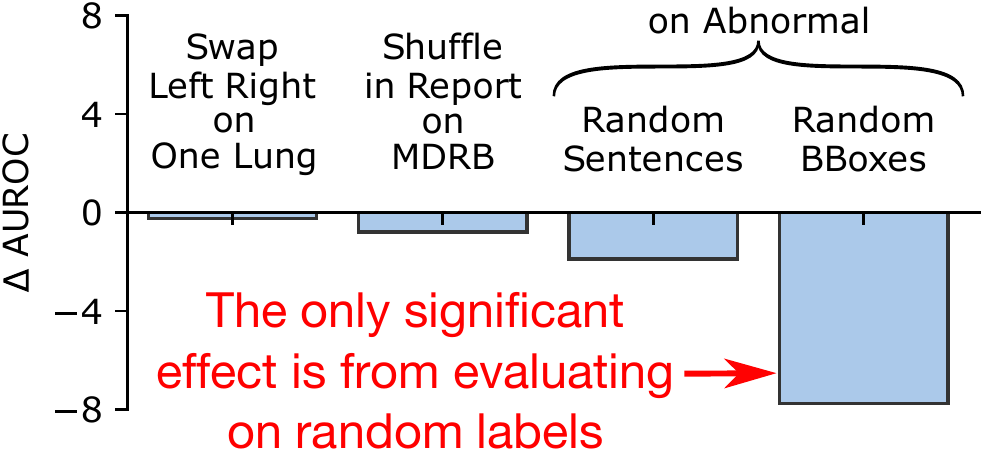}
    \caption{For each perturbation, we plot the change in localization performance (AUROC) of {\bf GLoRIA}.} 
    \label{fig:deltas_gloria}
\end{figure}

We first evaluate performance on the subsets described in Section \ref{sec:subsets}. 
This establishes a baseline 
with respect to which we can take differences observed under perturbations.
We report results in Table \ref{tab:subsets}. 
We observe that the model performs significantly worse on both the {\bf One Lung} and {\bf MDRB} subsets (which we view as ``harder'') in terms of AUROC and Average Precision, 
supporting this disaggregated evaluation.

Manual evaluation results of 3.1, 1.8, and 1.7 for recall, precision, and intuitiveness respectively indicate that GLoRIA produces unintuitive heatmaps that have poor precision and middling recall.
Because GLoria was trained on the CheXpert dataset and we perform these evaluations on ImaGenome, the change in dataset may be one cause of poor performance; in Section \ref{sec:improvements} we 
report how retraining on the ImaGenome dataset affects these scores.

\begin{table}[]
\small
    \centering
    \resizebox{\columnwidth}{!}{
    \begin{tabular}{l l l l l}
    \hline
        Subset & AUROC & Avg. P & IOU@5/10/30\% \\
        \hline
Abnormal & 69.51 & 48.29 & 4.10/7.25/19.05 \\
One Lung & 65.48 & 38.68 & 4.43/8.05/20.54 \\
MDRB & 65.01 & 36.96 & 3.56/6.37/16.92 \\
        \hline
    \end{tabular}
    }
    \caption{GLoRIA Localization performance on subsets.}
    \label{tab:subsets}
\end{table}

To measure the 
sensitivity of model attention to changes in the text, we report \textbf{differences in localization performance} in Figure \ref{fig:deltas_gloria}.
Specifically, this is the difference in model performance ($\Delta$AUROC) achieved using (a) the original (unperturbed) sentences, and, (b) sentences perturbed as described in Section \ref{sec:perturbations}. 
We show results for each perturbation on the subsets they should most effect (Section \ref{sec:subsets}), leaving the full results for the appendix (Figure \ref{fig:deltas_retrained_full}).

The only real decrease in performance observed is under the {\bf Random BBoxes} perturbation, which entails swapping out the target bounding box for an instance with one associated with some \emph{other instance (image)}.
Performance decreasing here (and not for text perturbations) is consistent with the hypothesis that the attention map primarily reflects the image modality, but not the text.
This is further supported by the observation that the model pays little mind to clear positional cue words such as ``left'' and ``right'' when constructing the attention map; witness the negligible drop in performance under the {\bf Swap Left Right} perturbation.
Finally, swapping in other sentences (even from different reports) yields almost no performance difference.


\section{Can We Improve Alignments?}\label{sec:improvements}

The above results indicate that 
image attention is unintuitive and less sensitive to the text modality than might be expected.
Next we propose simple methods to try to improve image/text alignments. 

\subsection{Models}\label{sec:models}

All models build on the GLoRIA architecture except the baseline \textbf{UNITER}, for which we perform no modifications except to re-train from scratch on the MIMIC-CXR/Chest ImaGenome dataset.\footnote{We re-train from scratch because: (1) Unlike in the original model, we are not feeding in features from Fast-RCNN, but instead using flattened pixels from a bounding box, and; (2) We would like a fair comparison to the GLoRIA variants which are also re-trained from scratch.}
In the results, \textbf{GLoRIA} refers to weights fit using the CheXpert dataset, released by \cite{Huang_2021_ICCV}.
We do not have access to the reports associated with this dataset 
so we do not use it for training or evaluation, but we do make comparisons to the original (released) \textbf{GLoRIA} model trained on it.

We also retrain our own {\bf GLoRIA} model on the MIMIC-CXR/ImaGenome dataset; we call this \textbf{GLoRIA Retrained}.
While the two datasets are similar in size and content, CheXpert has many more positive cases of conditions than MIMIC-CXR/ImaGenome (8.86\% of CheXpert images are labeled as having ``No Findings''; in the ImaGenome dataset, reports associated with 21.80\% of train images do not contain a sentence labeled ``abnormal'').
Given this difference in the number of positive cases, we train a \textbf{Retrained+Abnormal} model variant on the subset of MIMIC-CXR/ImaGenome sentence/image pairs featuring an ``abnormal'' sentence. 

We also train models in which we 
adopt masking strategies 
intended to improve localization, hypothesizing that this might prevent over-reliance on text artifacts that might allow the model to ignore text that localizes.
Our \textbf{Retrained+Word Masking} model randomly replaces words in the input with {\tt [MASK]} tokens during training with 30\% probability.\footnote{We choose the high value of 30\% here because without allowing hyperparameter tuning of this probability, we would like to see a significant impact when comparing to the baseline.}
For our \textbf{Retrained+Clinical Masking} model, we randomly swap clinical entity spans found using a {\tt SciSpaCy} entity linker \cite{neumann-etal-2019-scispacy} for {\tt [MASK]} tokens with 50\% probability.
Many sentences in a report will not refer to any particular region in the image.
We therefore propose the \textbf{Retrained+``No Attn'' Token} model, which concatenates a special ``No Attn'' token parameter vector to the set of local image embeddings just before attention is induced.
This allows the model to attend to this special vector, rather than any of the local image embeddings, effectively indicating that there is no good match.

We also 
consider a setting in which we assume a small 
amount of supervision (annotations linking image regions to texts).
We finetune a model to produce high attention on the annotated regions of interest, i.e., we supervise attention.
We employ an alignment loss
$\mathcal{L}_{\mathrm{alignment}}(s, \ell) = \sum_p s_p \ell_p$
using the pixel-wise scores $s$ derived from the attention\footnote{In this case, we also renormalize again after upsampling so the pixel scores to sum to 1.} and the segmentation labels $\ell$ (Section \ref{sec:data_and_metrics}).
We train on a batch of 30 examples for up to 500 steps with early stopping on an additional 30-example validation set using a patience of 25 steps.
This might be viewed as ``few-shot alignment'', where we use a small number of annotated examples to try to make the model more interpretable by improving image and text alignments.

Finally, as a point of reference we train \textbf{Retrained+Rand Sents} in the same style as the \textbf{Retrained} model except that all sentences are replaced with \emph{random} sentences.
This deprives the model of any meaningful training signal, which otherwise comes entirely through the pairing of images and texts.
This variant 
provides a baseline to help contextualize results.
For all models, we use early stopping with a patience of 10 epochs.\footnote{
For all models we report results on the last epoch before the early stopping condition is reached.}

\subsection{Results and Discussion}\label{sec:results_and_discussion}

\subsubsection{Localization Metrics}\label{sec:localization}

\begin{table}[]
    \centering
    \begin{tabular}{l l l}
    \hline
        Model & AUROC & Avg. P \\
        \hline
UNITER$^{\bigstar}$ & \textbf{84.92} & \textbf{68.57} \\
GLoRIA Retrained & 55.84 & 41.22 \\
\hline
+Word Masking & 61.44 & 44.69 \\
+Clinical Masking & 54.67 & 40.61 \\
+``No Attn'' Token & 57.00 & 41.80 \\
\hline
+Abnormal & 55.89 & 43.42 \\
+30-shot Finetuned & \textbf{63.90} & \textbf{52.80} \\
\hline
+Rand Sents & 38.88 & 30.55 \\
        \hline
    \end{tabular}
    \caption{Localization performance for each retrained model. $\bigstar$ UNITER here is not comparable because it uses ground truth bounding boxes as input. (Full results in Table \ref{tab:retrained_localization_performance_subsets}.)} 
    \label{tab:retrained_localization_performance}
\end{table}

Table \ref{tab:retrained_localization_performance} might 
seem to imply that \textbf{UNITER} performs best. However, we emphasize that this is not comparable to other models because, as discussed in \ref{sec:uniter}, \textbf{UNITER}'s attention is defined over ground truth anatomical bounding boxes (rather than the entire image), of which the sentence bounding boxes are a subset; this dramatically inflates AUROC and average precision scores.
(We have included \textbf{UNITER} despite this for completeness.)

Finetuning on a small set of ground truth bounding boxes (\textbf{+30-shot Finetuned}) substantially improves performance. 
Of the remaining (not explicitly supervised) approaches, \textbf{+Word Masking} fares best.
This masking may serve a regularization function similar to dropout \cite{JMLR:v15:srivastava14a}. 
Counter-intuitively, \textbf{+Clinical Masking} performs slightly worse than \textbf{Retrained}.
Perhaps clinical masking blinds too much key information.

The \textbf{+``No Attn'' Token} model also performs comparatively well, 
suggesting that allowing the model to not attend to any particular part of the image does increase performance.

To address a concern that the bounding boxes used as labels are bigger than the region of interest (Section \ref{sec:alignment_accuracy}), we try to improve our measure of precision by re-evaluating without some of the larger bounding boxes and find an overall drop in precision but with similar trends in relative model performance (see Section \ref{sec:dropping_large_bboxes}).

\subsubsection{Post-hoc Evaluation}

\begin{figure}
    \centering
    \includegraphics[scale=.5]{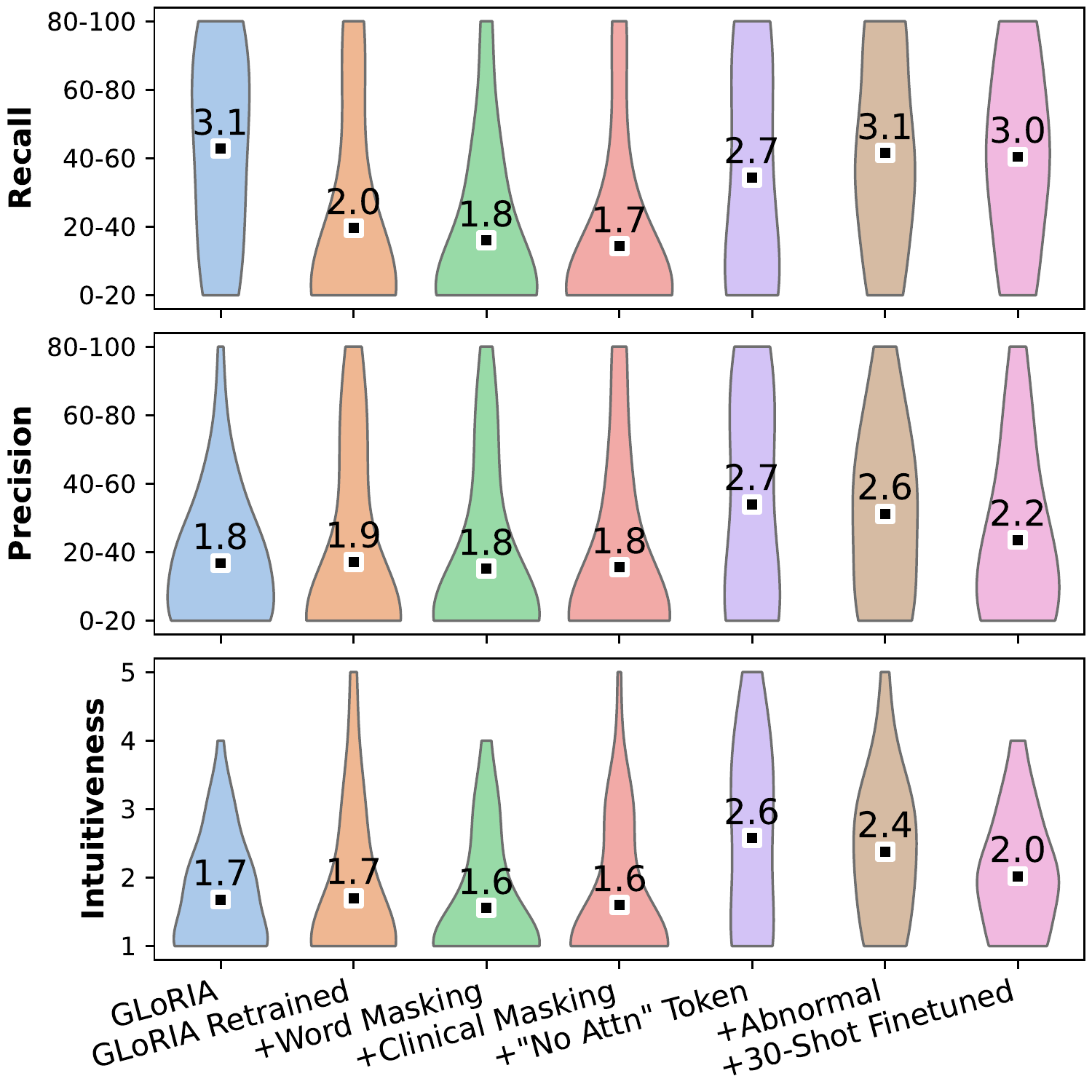}
    \caption{Annotations on Retrained Variants. We report means over 50 annotations. Recall and Precision scores 1, ..., 5 correspond to bins of 0-20, ..., 80-100 \%. UNITER is not included in this because the attention over the bounding boxes is very unintuitive and different from the other models' attention.}
    \label{fig:retrained_anns}
\end{figure}

The annotation results (Figure \ref{fig:retrained_anns}) of recall, precision, and intuitiveness are perhaps more revealing and do not necessarily align with our automatic metrics.\footnote{We do not include UNITER in this because the attention over the bounding boxes is very unintuitive and different from the other models’ attention (See Appendix Figure \ref{fig:uniter_attn}).} 
This is likely a product of the limitations of the ImaGenome bounding boxes.
The \textbf{+``No Attn'' Token} model scored highest in terms of intuitiveness and precision, which is promising given that unlike the \textbf{+Abnormal} and \textbf{+30-shot Finetuned} models, this model does not 
require any additional training information (i.e., indications of training sample abnormalities, or ground truth bboxes).
A simple modification to the architecture that allows it to pass on aligning a given text to the image yields a stark increase in performance with respect to the baseline \textbf{Retrained} model.
The \textbf{Retrained} model 
performs about the same as \textbf{GLoRIA} in terms of precision and intuitiveness, although it incurs a significant drop in recall.

The \textbf{+30-shot Finetuned} model uses the bounding boxes as ground truth, but these are somewhat noisy. 
Better annotations of the regions of interest 
might improve intuitiveness further. 
When performing annotations, the radiologist also noticed that a large percentage of sentences in reports do not refer to anything focal, which indicates the necessity of looking at the subsets from Section \ref{sec:subsets}---all of which should have more focal sentences---especially when it comes to the perturbations.
This also may help explain the superior performance of the \textbf{+``No Attn'' Token} model which explicitly handles these cases.

\begin{figure}
    \centering
    \includegraphics[scale=.75]{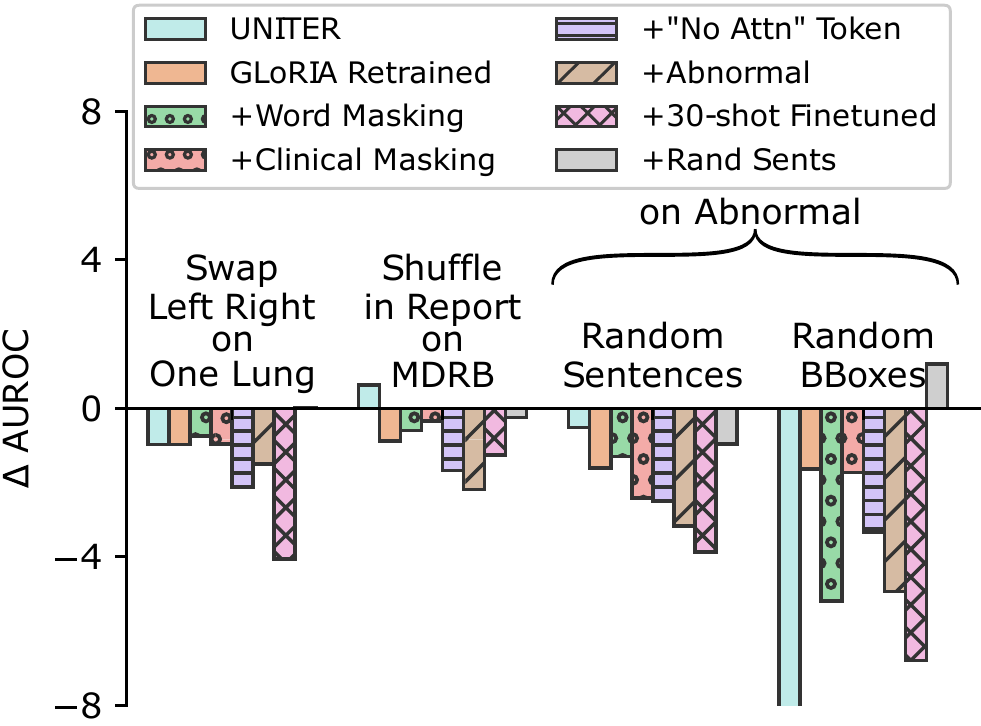}
    \caption{For each perturbation, we plot the change in localization performance (measured by AUROC), for each of the models we retrain from scratch on the respective subsets. Here, UNITER is effected most by the Random BBoxes perturbations because it uses the original ground truth as input.}
    \label{fig:deltas_retrained}
\end{figure}

\subsubsection{Perturbation Results}

We next perform the perturbations 
introduced above (and assessed on GLoRIA) to the proposed variants to assess sensitivity to input texts (full results in Figure \ref{fig:deltas_retrained_full} of the appendix).
We observe that \textbf{+30-shot Finetuned}, \textbf{+``No Attn'' Token}, and \textbf{+Abnormal}, in that order, are  most affected when swapping left and right.
These three models are also the most affected by shuffling bounding boxes within a report or swapping in a random sentence from the rest of the dataset, although for these perturbations, the {\bf +Abnormal} model is more sensitive than the {\bf +``No Attn'' Token}.


The \textbf{Random BBoxes} perturbation serves mostly as a 
reference measure of how variable model scores can 
be when swapping in entirely wrong bounding boxes.
But it also seems to suggest that for models affected more by this, the attention is more focused on precision.
This indicates that besides UNITER, the \textbf{+30-shot Finetuned}, \textbf{+Word Masking}, \textbf{+Abnormal}, and \textbf{+``No Attn'' Token}, in that order, are the most precise; 
this is in line with the average precision scores in Table \ref{tab:retrained_localization_performance} and the entropy scores in the appendix (Table \ref{tab:attn_entropy}).

Taken together these perturbation results 
suggest that \textbf{+``No Attn'' Token}, \textbf{+Abnormal}, and \textbf{+30-shot Fine-tuned} are the models most intuitively sensitive to text.
However, they remain less intuitive than they would ideally be.\footnote{We discuss results for experiments in which we swap conditions in synthetic sentences in Appendix (\ref{sec:deltas_app}); these are inconclusive.}

\subsubsection{Contrastive Accuracy}

Table \ref{tab:candidate_selection} reports the accuracy of each model 
with respect to identifying the correct sentence from two candidates for a given image.
These results indicate that performing comparatively well at identifying the correct sentence does not necessarily correlate with intuitiveness or textual sensitivity, i.e., being able to discriminate between \emph{sentences} given an image does not imply an ability to accurately \emph{localize} within an image, given a sentence.
In particular, {\bf +Word Masking} performs best here, though we saw above that it is relatively unintuitive and its localization is somewhat invariant to perturbations. 
Further, the three best models in terms of textual sensitivity have relatively poor performance (with the possible exception of the \textbf{+Abnormal} variant).

\begin{table}[]
    \footnotesize
    \centering
    \begin{tabular}{l l l l l}
    \hline
        & \multicolumn{2}{c}{{\bf All}} & \multicolumn{2}{c}{{\bf Abnormal}} \\
        Model & local & global & local & global \\
        \hline
UNITER & - & 67.2 & - & 70.7 \\
GLoRIA & 55.2 & 70.3 & 43.3 & 77.0 \\
GLoRIA Retrained & 70.2 & \textbf{82.9} & 63.8 & 86.4 \\
+Word Masking & \textbf{78.9} & 81.6 & \textbf{80.3} & \textbf{86.5} \\
+Clinical Masking & 68.5 & 81.5 & 65.4 & 84.4 \\
+``No Attn'' Token & 67.3 & 81.9 & 61.8 & 85.0 \\
+Abnormal & 72.1 & 76.6 & 73.1 & 84.1 \\
+30-shot Finetuned & 67.2 & 79.6 & 61.0 & 83.6 \\
+Rand Sents & 51.4 & 51.3 & 44.8 & 60.6 \\
\hline
    \end{tabular}
    \vspace{-.5em}
    \caption{Average accuracies with respect to discriminating between the sentence actually associated with an image and a sentence randomly sampled from the dataset. (See Appendix Table \ref{tab:candidate_selection_subsets} for results on subsets.) Global and local refer to using only global or local embeddings for computing similarity.} 
    \label{tab:candidate_selection}
\end{table}

\subsubsection{Metric Correlations}

\begin{figure}
    \centering
    \includegraphics[scale=.67]{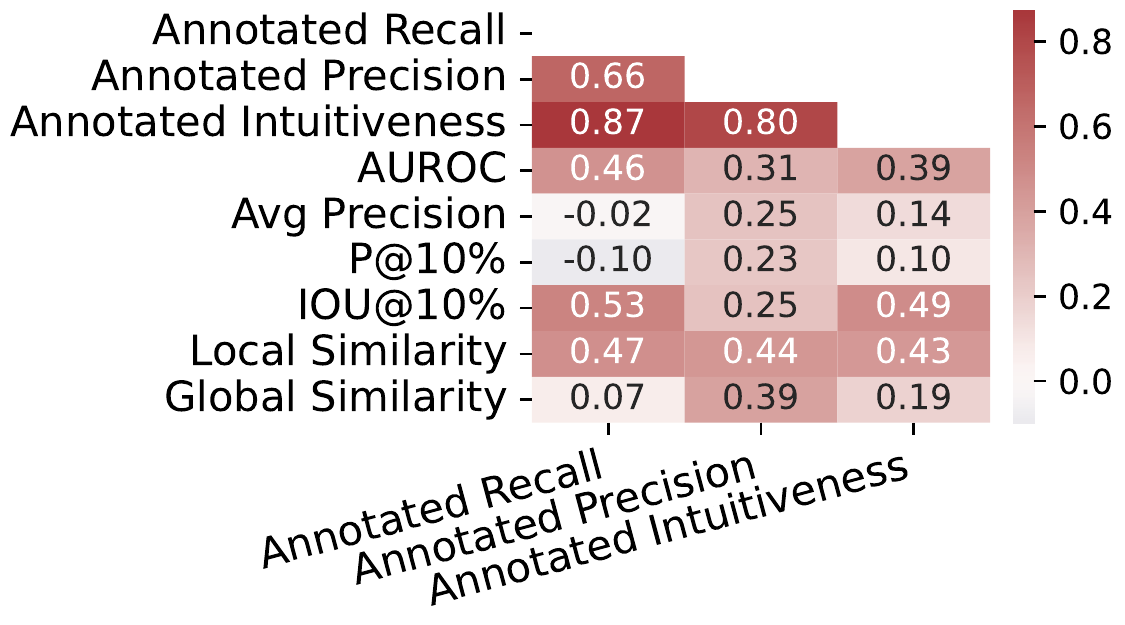}
    \caption{\textbf{Correlations} between metrics and annotations.}
    \label{fig:correlations}
\end{figure}

To 
quantify the relationships between scores, we report correlations between them  
across instances for \textbf{+``No Attn'' Token} (the best model in terms of manually judged intuitiveness) in Figure \ref{fig:correlations}.
Of the automatic metrics, IOU@10\% has the strongest correlation with annotated intuitiveness. 
Avg. Precision and Precision at 10\% have almost no correlation with intuitiveness and relatively weak correlation with annotated precision.
We also show correlation with local and global similarity between two positive pairs.\footnote{Because we only look at positive pairs, higher similarity scores are better.}
Though the local similarity of positive pairs is somewhat correlated with each of the annotation ratings, the global similarity is only (weakly) correlated with annotated precision.

The ``No Attn'' score, which is what we use to refer to the attention score for the added ``No Attn'' token, has some interesting Pearson correlations. 
Unfortunately, 
its relationship with annotations 
is complicated by our user interface. 
Often the ``No Attn'' score (which we display in the corner of the image) will either be unnoticeable or it will saturate the heatmap, resulting in the radiologist assigning low scores (1s) for an instance.
Therefore, we note that some negative correlations with annotations (Figure \ref{fig:corr_na}) may 
mostly reflect how the X-ray heatmap is displayed to the user.
However, a -.30 correlation with IOU@10\% and a -.47 correlation with whether 
an image contains an abnormality are significant. 
This demonstrates the potential for this score to identify situations where the model should abstain from displaying a heatmap altogether because either there is nothing abnormal to highlight or the model is not confident in its heatmap.

\subsubsection{Qualitative Analysis}

We note some interesting qualitative behavior discovered during the annotation that may also support the use of this ``No Attn'' architecture and score.
When many of the models are incorrect, they tend to highlight image edges or corners.
We hypothesize this occurs because the model attempts to find a static part of the image---one that is similar across most instances---on which to attend.
This behavior is misleading and not quantifiable.
The ``No Attn'' Token offers an alternative to this behavior, providing a means for the model to pass on inducing a heatmap altogether when appropriate. 

\begin{figure*}
    \centering
    \includegraphics[scale=.35]{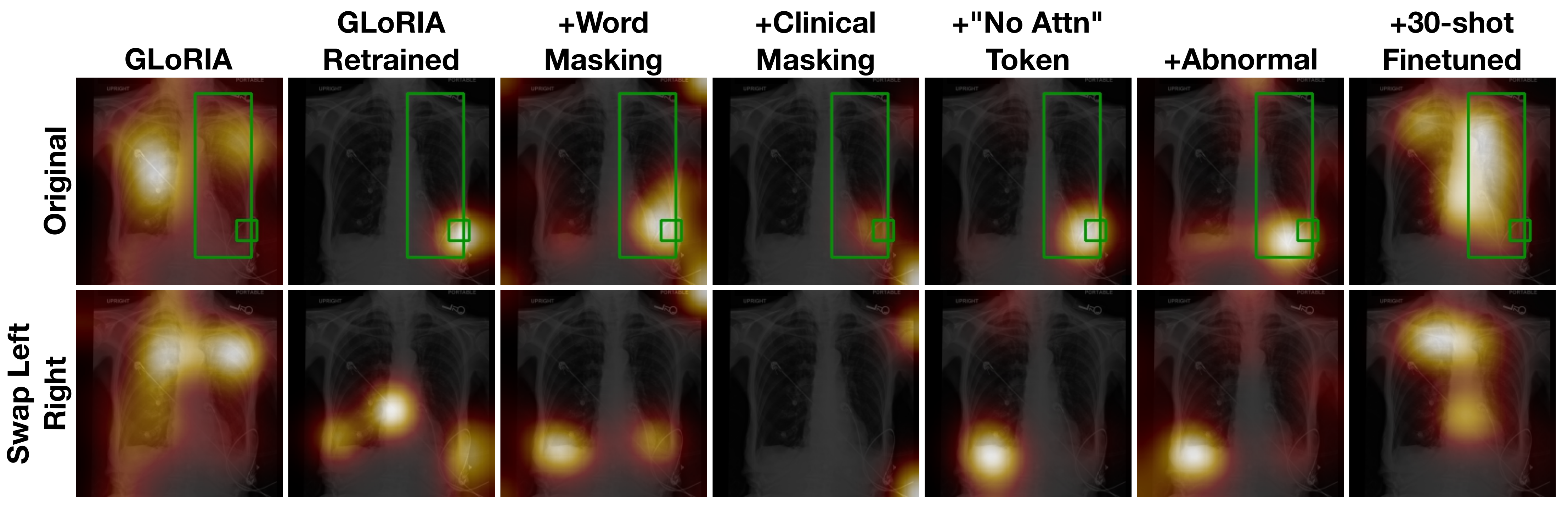}
    \caption{Attention for ``Blunting of the left costophrenic angle suggests small effusion.'' (top), and perturbed version (bottom).}
    \label{fig:qualitative}
\end{figure*}

We conclude with a qualitative impression of localization performance.
Figure \ref{fig:qualitative} shows model attention distributions for a (cherry-picked) instance and the accompanying {\bf Swap Left Right} perturbation.
This example was selected specifically to illustrate how models can fail to behave intuitively. 
In this example,
the correct region of interest for the original prompt lies mostly centered on the small box, and the large box (corresponding to the left lung) is somewhat misleading as it covers more than the strict region of interest.
This example demonstrates that though the anatomical locations discussed in the prompt are correctly highlighted by the bounding boxes, the \textit{region of interest} is not always directly on those anatomical locations.

With the original prompt, {\bf GLoRIA} yields a high-entropy map, {\bf GLoRIA Retrained} and the {\bf +Masking}, {\bf +``No Attn'' Token}, and {\bf+Abnormal} are centered roughly correctly (some more intuitive than others), and finally, \textbf{+30-shot Finetuned} almost fully highlights the large box (even though this is not strictly the correct region of interest) and almost entirely ignores the small box (the real region of interest).
The perturbation of swapping out ``left'' with ``right'' changes all of the models' heatmaps to varying degrees and with varying intuitiveness.
In this example, the most intuitive heatmaps after the perturbation are given by the \textbf{+``No Attn'' Token} and \textbf{+Abnormal} models, whereas other models still show significant emphasis on the original region and/or show emphasis on unintuitive and entirely irrelevant regions.

\vspace{0.3em}
\noindent{\bf Summary of key findings} 
Existing multimodal pretraining schemes beget models that accurately select the text that matches a given image (Table \ref{tab:candidate_selection}), and yield attention distributions that at least somewhat depend on the text.
But these models are not found intuitive (Table \ref{fig:retrained_anns}) and perturbing texts does not cause not consistently yield changes in the attention patterns that one would expect (Figure \ref{fig:deltas_retrained}).
Simple changes to pre-training may improve this behavior. 
Specifically, adding the ability of the model to \emph{not attend to any particular part of the image} may result in models that produce attention patterns which are more intuitive (Figure \ref{fig:retrained_anns}) \textit{and} more reflective of input texts (Figure \ref{fig:deltas_retrained}), although this may slightly harm performance on the pre-training task itself (Table \ref{tab:candidate_selection}).
\section{Related Work}

Work on multi-modal representation learning for medical data has proposed soft aligning modalities, but has focussed quantitative evaluation on the resultant performance that learned representations afford on downstream tasks \cite{Ji2021ImprovingJL, Liao2021MultimodalRL, Huang_2021_ICCV}. Model interpretability is often suggested using only qualitative examples; our work aims to close this gap.

A line of work in NLP evaluates the interpretability of neural attention mechanisms \cite{Jain2019AttentionIN, wiegreffe-pinter-2019-attention,serrano2019attention}.
Elsewhere, work at the intersection of computer vision and radiology has critically evaluated use of saliency maps over images \cite{arun2021assessing, rajpurkar2018deep}. 

Recent work has sought to improve the ability of these models to identify fine-grained alignments 
via supervised attention \cite{Kervadec2020WeakSH, Sood2021MultimodalIO}, but have focused on downstream task performance.
This differs from our focus on evaluating and improving localization itself, especially within the medical domain.
We also do not assume access to large amounts of supervision, which is commonly lacking in this domain.




\section{Conclusions}

We evaluated existing state-of-the-art unsupervised multimodal representation learning models for EHRs in terms of inducing fine-grained alignments between image regions and text.
We found that the resultant heatmaps are often unintuitive and invariant to perturbations to the text that ought to change them substantially.

We evaluated a number of methods aimed at improving this behavior, finding that: (1) allowing the model to refrain from attending to the image, and; (2) finetuning the model on a small set of labels for interpretable heatmaps substantially improves performance.
We hope that this effort motivates more work addressing the interpreteability of multi-modal encoders for healthcare.


\vspace{-5pt}
\section*{Limitations}
\vspace{-5pt}

This is a first attempt to investigate the interpretability of pre-trained multi-modals models for medical imaging data, and as such our work has important limitations.
ImaGenome only annotates anatomical locations for each sentence and bounding boxes for each anatomical location; these may not correspond directly to regions of interest.
In addition, these extracted anatomical locations highlight many levels of the hierarchy, so if a specific part of the lung is mentioned, the whole lung's bounding box may still be included.
The annotations we collected for evaluation also have important limitations to consider, namely that we only used one radiologist annotator and only annotated 50 instances.

Finally, we did not try a more fine-grained UNITER model with more and smaller bounding boxes that form a grid (to avoid an object detector), primarily because this would incur a significantly higher computational cost due to the number of image vectors; future work might explore this option.

\vspace{-5pt}
\section*{Ethics}
\vspace{-5pt}

There are significant risks associated with incorrectly interpreting models in the medical domain.
Our aim in this work is to highlight gaps in the current technology and suggest avenues for future research and \textit{not} to provide deployable models.
These models are not ready for use in the field because they may mislead users about the underlying reasons for predictions and incorrectly inform resulting decisions.
We hope this work facilitates advances in the interpretability of these models so that they may eventually provide meaningful guidance to radiologists.
The MIMIC-CXR dataset used was licensed via the PhysioNet Credentialed Health Data License 1.5.0, and we properly comply with the PhysioNet Credentialed Health Data Use Agreement 1.5.0.

\vspace{-2pt}
\section*{Acknowledgements}
\vspace{-3pt}

We acknowledge partial funding for this work by National Library of Medicine of the National Institutes of Health under award numbers R01LM013772 and R01LM013891. The content is solely the responsibility of the authors and does not necessarily represent the official views of the National Institutes of Health.
The work was also supported in part by the National Science Foundation (NSF) grant 1901117.
\bibliography{anthology,custom}

\begin{thebibliography}{29}
\expandafter\ifx\csname natexlab\endcsname\relax\def\natexlab#1{#1}\fi

\bibitem[{Alsentzer et~al.(2019)Alsentzer, Murphy, Boag, Weng, Jin, Naumann,
  and McDermott}]{Alsentzer2019PubliclyAC}
Emily Alsentzer, John~R. Murphy, Willie Boag, Wei-Hung Weng, Di~Jin, Tristan
  Naumann, and Matthew B.~A. McDermott. 2019.
\newblock Publicly available clinical bert embeddings.
\newblock \emph{ArXiv}, abs/1904.03323.

\bibitem[{Arun et~al.(2021)Arun, Gaw, Singh, Chang, Aggarwal, Chen, Hoebel,
  Gupta, Patel, Gidwani et~al.}]{arun2021assessing}
Nishanth Arun, Nathan Gaw, Praveer Singh, Ken Chang, Mehak Aggarwal, Bryan
  Chen, Katharina Hoebel, Sharut Gupta, Jay Patel, Mishka Gidwani, et~al. 2021.
\newblock Assessing the trustworthiness of saliency maps for localizing
  abnormalities in medical imaging.
\newblock \emph{Radiology: Artificial Intelligence}, 3(6):e200267.

\bibitem[{Bahdanau et~al.(2014)Bahdanau, Cho, and Bengio}]{bahdanau2014neural}
Dzmitry Bahdanau, Kyunghyun Cho, and Yoshua Bengio. 2014.
\newblock Neural machine translation by jointly learning to align and
  translate.
\newblock \emph{arXiv preprint arXiv:1409.0473}.

\bibitem[{Chauhan et~al.(2020)Chauhan, Liao, Wells, Andreas, Wang, Berkowitz,
  Horng, Szolovits, and Golland}]{Chauhan2020JointMO}
Geeticka Chauhan, Ruizhi Liao, William~M. Wells, Jacob Andreas, Xin Wang,
  Seth~J. Berkowitz, Steven Horng, Peter Szolovits, and Polina Golland. 2020.
\newblock Joint modeling of chest radiographs and radiology reports for
  pulmonary edema assessment.
\newblock \emph{Medical image computing and computer-assisted intervention :
  MICCAI ... International Conference on Medical Image Computing and
  Computer-Assisted Intervention}, 12262:529--539.

\bibitem[{Chen et~al.(2020)Chen, Li, Yu, Kholy, Ahmed, Gan, Cheng, and
  Liu}]{Chen2020UNITERUI}
Yen-Chun Chen, Linjie Li, Licheng Yu, Ahmed~El Kholy, Faisal Ahmed, Zhe Gan,
  Yu~Cheng, and Jingjing Liu. 2020.
\newblock Uniter: Universal image-text representation learning.
\newblock In \emph{ECCV}.

\bibitem[{Gan et~al.(2020)Gan, Chen, Li, Zhu, Cheng, and Liu}]{gan2020large}
Zhe Gan, Yen-Chun Chen, Linjie Li, Chen Zhu, Yu~Cheng, and Jingjing Liu. 2020.
\newblock Large-scale adversarial training for vision-and-language
  representation learning.
\newblock In \emph{NeurIPS}.

\bibitem[{He et~al.(2016)He, Zhang, Ren, and Sun}]{He2016DeepRL}
Kaiming He, X.~Zhang, Shaoqing Ren, and Jian Sun. 2016.
\newblock Deep residual learning for image recognition.
\newblock \emph{2016 IEEE Conference on Computer Vision and Pattern Recognition
  (CVPR)}, pages 770--778.

\bibitem[{Huang et~al.(2021)Huang, Shen, Lungren, and Yeung}]{Huang_2021_ICCV}
Shih-Cheng Huang, Liyue Shen, Matthew~P. Lungren, and Serena Yeung. 2021.
\newblock Gloria: A multimodal global-local representation learning framework
  for label-efficient medical image recognition.
\newblock In \emph{Proceedings of the IEEE/CVF International Conference on
  Computer Vision (ICCV)}, pages 3942--3951.

\bibitem[{Huang et~al.(2020)Huang, Zeng, Liu, Fu, and
  Fu}]{Huang2020PixelBERTAI}
Zhicheng Huang, Zhaoyang Zeng, Bei Liu, Dongmei Fu, and Jianlong Fu. 2020.
\newblock Pixel-bert: Aligning image pixels with text by deep multi-modal
  transformers.
\newblock \emph{ArXiv}, abs/2004.00849.

\bibitem[{Jain and Wallace(2019)}]{Jain2019AttentionIN}
Sarthak Jain and Byron~C. Wallace. 2019.
\newblock Attention is not explanation.
\newblock In \emph{NAACL}.

\bibitem[{Ji et~al.(2021)Ji, Shaikh, Moukheiber, Srihari, Peng, and
  Gao}]{Ji2021ImprovingJL}
Zhanghexuan Ji, Mohammad~Abuzar Shaikh, Dana Moukheiber, Sargur~N. Srihari,
  Yifan Peng, and Mingchen Gao. 2021.
\newblock Improving joint learning of chest x-ray and radiology report by word
  region alignment.
\newblock In \emph{MLMI@MICCAI}.

\bibitem[{Johnson et~al.(2019{\natexlab{a}})Johnson, Pollard, Berkowitz,
  Greenbaum, Lungren, ying Deng, Mark, and Horng}]{Johnson2019MIMICCXRAD}
Alistair E.~W. Johnson, Tom~J. Pollard, Seth~J. Berkowitz, Nathaniel~R.
  Greenbaum, Matthew~P. Lungren, Chih ying Deng, Roger~G. Mark, and Steven
  Horng. 2019{\natexlab{a}}.
\newblock Mimic-cxr, a de-identified publicly available database of chest
  radiographs with free-text reports.
\newblock \emph{Scientific Data}, 6.

\bibitem[{Johnson et~al.(2019{\natexlab{b}})Johnson, Pollard, Berkowitz,
  Greenbaum, Lungren, ying Deng, Mark, and Horng}]{Johnson2019MIMICCXRAL}
Alistair E.~W. Johnson, Tom~J. Pollard, Seth~J. Berkowitz, Nathaniel~R.
  Greenbaum, Matthew~P. Lungren, Chih ying Deng, Roger~G. Mark, and Steven
  Horng. 2019{\natexlab{b}}.
\newblock Mimic-cxr: A large publicly available database of labeled chest
  radiographs.
\newblock \emph{ArXiv}, abs/1901.07042.

\bibitem[{Kervadec et~al.(2020)Kervadec, Antipov, Baccouche, and
  Wolf}]{Kervadec2020WeakSH}
Corentin Kervadec, Grigory Antipov, Moez Baccouche, and Christian Wolf. 2020.
\newblock Weak supervision helps emergence of word-object alignment and
  improves vision-language tasks.
\newblock \emph{ArXiv}, abs/1912.03063.

\bibitem[{Li et~al.(2019)Li, Yatskar, Yin, Hsieh, and
  Chang}]{Li2019VisualBERTAS}
Liunian~Harold Li, Mark Yatskar, Da~Yin, Cho-Jui Hsieh, and Kai-Wei Chang.
  2019.
\newblock Visualbert: A simple and performant baseline for vision and language.
\newblock \emph{ArXiv}, abs/1908.03557.

\bibitem[{Li et~al.(2020)Li, Wang, and Luo}]{Li2020ACO}
Yikuan Li, Hanyin Wang, and Yuan Luo. 2020.
\newblock A comparison of pre-trained vision-and-language models for multimodal
  representation learning across medical images and reports.
\newblock \emph{2020 IEEE International Conference on Bioinformatics and
  Biomedicine (BIBM)}, pages 1999--2004.

\bibitem[{Liao et~al.(2021)Liao, Moyer, Cha, Quigley, Berkowitz, Horng,
  Golland, and Wells}]{Liao2021MultimodalRL}
Ruizhi Liao, Daniel Moyer, Miriam Cha, Keegan Quigley, Seth~J. Berkowitz,
  Steven Horng, Polina Golland, and William~M. Wells. 2021.
\newblock Multimodal representation learning via maximization of local mutual
  information.
\newblock In \emph{MICCAI}.

\bibitem[{Lin et~al.(2014)Lin, Maire, Belongie, Hays, Perona, Ramanan, Dollar,
  and Zitnick}]{lin2014microsoft}
Tsung-Yi Lin, Michael Maire, Serge Belongie, James Hays, Pietro Perona, Deva
  Ramanan, Piotr Dollar, and Larry Zitnick. 2014.
\newblock \href
  {https://www.microsoft.com/en-us/research/publication/microsoft-coco-common-objects-in-context/}
  {Microsoft coco: Common objects in context}.
\newblock In \emph{ECCV}. European Conference on Computer Vision.

\bibitem[{Neumann et~al.(2019)Neumann, King, Beltagy, and
  Ammar}]{neumann-etal-2019-scispacy}
Mark Neumann, Daniel King, Iz~Beltagy, and Waleed Ammar. 2019.
\newblock \href {https://doi.org/10.18653/v1/W19-5034} {{S}cispa{C}y: {F}ast
  and {R}obust {M}odels for {B}iomedical {N}atural {L}anguage {P}rocessing}.
\newblock In \emph{Proceedings of the 18th BioNLP Workshop and Shared Task},
  pages 319--327, Florence, Italy. Association for Computational Linguistics.

\bibitem[{Rajpurkar et~al.(2018)Rajpurkar, Irvin, Ball, Zhu, Yang, Mehta, Duan,
  Ding, Bagul, Langlotz et~al.}]{rajpurkar2018deep}
Pranav Rajpurkar, Jeremy Irvin, Robyn~L Ball, Kaylie Zhu, Brandon Yang, Hershel
  Mehta, Tony Duan, Daisy Ding, Aarti Bagul, Curtis~P Langlotz, et~al. 2018.
\newblock Deep learning for chest radiograph diagnosis: A retrospective
  comparison of the chexnext algorithm to practicing radiologists.
\newblock \emph{PLoS medicine}, 15(11):e1002686.

\bibitem[{Serrano and Smith(2019)}]{serrano2019attention}
Sofia Serrano and Noah~A Smith. 2019.
\newblock Is attention interpretable?
\newblock \emph{arXiv preprint arXiv:1906.03731}.

\bibitem[{Sood et~al.(2021)Sood, K{\"o}gel, Muller, Thomas, B{\^a}ce, and
  Bulling}]{Sood2021MultimodalIO}
Ekta Sood, Fabian K{\"o}gel, Philippe Muller, Dominike Thomas, Mihai B{\^a}ce,
  and Andreas Bulling. 2021.
\newblock Multimodal integration of human-like attention in visual question
  answering.
\newblock \emph{ArXiv}, abs/2109.13139.

\bibitem[{Srivastava et~al.(2014)Srivastava, Hinton, Krizhevsky, Sutskever, and
  Salakhutdinov}]{JMLR:v15:srivastava14a}
Nitish Srivastava, Geoffrey Hinton, Alex Krizhevsky, Ilya Sutskever, and Ruslan
  Salakhutdinov. 2014.
\newblock \href {http://jmlr.org/papers/v15/srivastava14a.html} {Dropout: A
  simple way to prevent neural networks from overfitting}.
\newblock \emph{Journal of Machine Learning Research}, 15(56):1929--1958.

\bibitem[{Su et~al.(2020)Su, Zhu, Cao, Li, Lu, Wei, and Dai}]{Su2020VLBERTPO}
Weijie Su, Xizhou Zhu, Yue Cao, Bin Li, Lewei Lu, Furu Wei, and Jifeng Dai.
  2020.
\newblock Vl-bert: Pre-training of generic visual-linguistic representations.
\newblock \emph{ArXiv}, abs/1908.08530.

\bibitem[{Tan and Bansal(2019)}]{Tan2019LXMERTLC}
Hao~Hao Tan and Mohit Bansal. 2019.
\newblock Lxmert: Learning cross-modality encoder representations from
  transformers.
\newblock In \emph{EMNLP}.

\bibitem[{Wang et~al.(2018)Wang, Peng, Lu, Lu, and Summers}]{Wang2018TieNetTE}
Xiaosong Wang, Yifan Peng, Le~Lu, Zhiyong Lu, and Ronald~M. Summers. 2018.
\newblock Tienet: Text-image embedding network for common thorax disease
  classification and reporting in chest x-rays.
\newblock \emph{2018 IEEE/CVF Conference on Computer Vision and Pattern
  Recognition}, pages 9049--9058.

\bibitem[{Wiegreffe and Pinter(2019)}]{wiegreffe-pinter-2019-attention}
Sarah Wiegreffe and Yuval Pinter. 2019.
\newblock \href {https://doi.org/10.18653/v1/D19-1002} {Attention is not not
  explanation}.
\newblock In \emph{Proceedings of the 2019 Conference on Empirical Methods in
  Natural Language Processing and the 9th International Joint Conference on
  Natural Language Processing (EMNLP-IJCNLP)}, pages 11--20, Hong Kong, China.
  Association for Computational Linguistics.

\bibitem[{Wu et~al.(2021)Wu, Agu, Lourentzou, Sharma, Paguio, Yao, Dee,
  Mitchell, Kashyap, Giovannini, Celi, and Moradi}]{Wu2021ChestID}
Joy~T. Wu, Nkechinyere~N. Agu, Ismini Lourentzou, Arjun Sharma,
  Joseph~Alexander Paguio, Jasper~Seth Yao, Edward~Christopher Dee, William
  Mitchell, Satyananda Kashyap, Andrea Giovannini, Leo~Anthony Celi, and Mehdi
  Moradi. 2021.
\newblock Chest imagenome dataset for clinical reasoning.
\newblock \emph{ArXiv}, abs/2108.00316.

\bibitem[{Zech et~al.(2018)Zech, Badgeley, Liu, Costa, Titano, and
  Oermann}]{10.1371/journal.pmed.1002683}
John~R. Zech, Marcus~A. Badgeley, Manway Liu, Anthony~B. Costa, Joseph~J.
  Titano, and Eric~Karl Oermann. 2018.
\newblock \href {https://doi.org/10.1371/journal.pmed.1002683} {Variable
  generalization performance of a deep learning model to detect pneumonia in
  chest radiographs: A cross-sectional study}.
\newblock \emph{PLOS Medicine}, 15(11):1--17.

\end{thebibliography}
\bibliographystyle{acl_natbib}

\appendix

\clearpage
\section{More details}\label{sec:appendix}

\subsection{Challenges Applying General Domain Multimodal Models to Medical Data}
\label{sec:general_domain_challenges}

Given recent progress made on open domain multi-modal models, e.g., UNITER \cite{Chen2020UNITERUI},  it is reasonable to ask whether we can simply apply such models and pre-training schemes to multi-modal medical data. 
However, a few key difficulties complicate straight-forward adaptation.

\textbf{Necessity of Object Detectors.} 
Many open domain models 
assume access to general \emph{object detectors} during pre-processing.
Such detectors are not readily available in the medical domain,
and training object detection models requires large-scale, high-quality annotations for many different phenomena and/or anatomical regions.
Further, one would need to collect such data 
for each domain in radiology (e.g., brain versus chest imaging).

In many multimodal models object detectors are used to produce bounding boxes, 
and are also tasked with inducing low-dimensional fixed-length vectors for significant regions, 
effectively taking care of region representation learning so that it need not be learned end-to-end.
Open domain models often expect tens of bounding boxes in an image, but even a coarse segmentation of 
images (e.g., into a 19x19 grid as in GLoRIA) yields many more bounding boxes than this, 
exacerbating the mismatch between pre-trained general object detectors and the medical domain when the former are initialized from open-domain checkpoints.


\textbf{Mismatch in Alignment Assumptions.} UNITER uses \textbf{optimal transport} to align image and text vectors, but this assumes that each object (or salient part within an image) can be reasonably aligned to a segment of the corresponding text. 
This makes sense in the case of the general domain data like COCO \cite{lin2014microsoft} because usually we expect most detected objects to be mentioned in the caption.
By contrast, in the medical domain 
we 
would expect that \emph{most} parts of the image are unrelated to \emph{any} portion of the corresponding text, 
and the task of the model is to identify salient regions of the text and match these with a particular image region.
This is especially true when not using an object detector to identify the interesting regions as preprocessing step.
The result of the optimal transport objective is that, averaging over tokens in the input text, each bounding box is equally important.
In Section \ref{sec:uniter}, we 
circumvent this problem by not using the optimal transport distribution itself (though that would be the natural choice), but instead using the attention mechanisms within UNITER.

Despite these obstacles to  re-purposing open domain multimodal models for this space, in Section \ref{sec:uniter-details} we describe how we 
modify UNITER to serve as a baseline for our analysis for completeness. 

\subsection{UNITER Details} \label{sec:uniter-details}

\begin{figure}
    \centering
    \includegraphics[scale=.6]{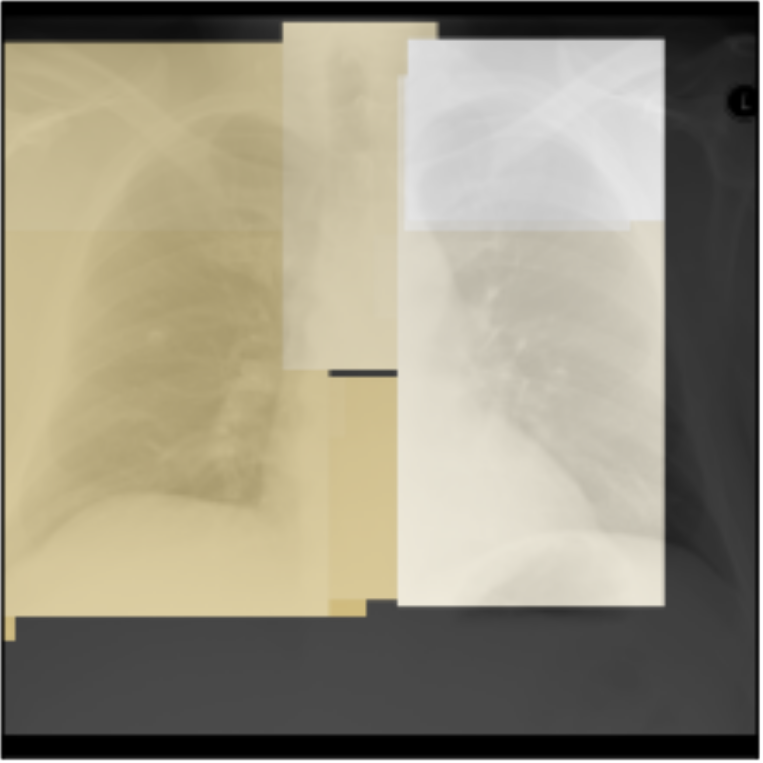}
    \caption{Here we show an example of what the UNITER attention over the ground truth bounding boxes looks like. It is easy to see why this attn usually has high localization with respect to the bounding boxes but also remains very unintuitive. It is still unclear if the UNITER model learns how to localize at all.}
    \label{fig:uniter_attn}
\end{figure}

We use all reference anatomical bounding boxes 
available in the ImaGenome dataset.
We 
reshape each bounding box to 45 $\times$ 45 pixels (enforcing fixed length), and then flatten and zero-pad the resultant vectors to be of length 2048 (the dimension UNITER expects). We train UNITER with a batch size of 4096 and 5 gradient accumulation steps on ImaGenome for 200k training steps. 

For saliency we compute the mean attention over all 144 heads (12 layers $\times$ 12 heads) to produce pixel-wise scores \cite{gan2020large}.
We take the mean of the attention when querying the text over the image, and when querying the image over the text; we normalize the resultant scores and treat these as analogous to $a_{ij}$, i.e., the saliency score relating the two modalities.
We note that the absolute overlap scores between UNITER attention (just defined) and bounding boxes will be relatively high given that the UNITER attention is defined over the ground truth bounding boxes for all anatomical locations, and the bounding boxes used to evaluate the attention for a particular sentence are a subset of these same input bounding boxes.
This also means we cannot use this approach in practice in the unsupervised setting in which we operate.

\subsection{Synthetic Sentences}\label{sec:synthetic_sents}

In Section \ref{sec:results_app}, we include results involving synthetic sentences, which we describe here.
To facilitate controlled experiments involving swapping out conditions --- Section \ref{sec:perturbations}, {\bf Synthetic+Swapped Conditions} --- we also adopt a strategy for creating \textbf{synthetic sentences} using the labels from ImaGenome \cite{Wu2021ChestID}, and test our models on these sentences as well.
Specifically, we construct these sentences using a set of rules to translate the condition and positive/negative context annotations and the anatomical names for the corresponding bounding boxes into natural language, as described in Table \ref{sec:synthetic_sents}.\footnote{We present examples in the Appendix (Table \ref{tab:synthetic_sents_examples}).}

\subsubsection{Rules for Creating Synthetic Sentences}

    \resizebox{\columnwidth}{!}{
    \begin{tabular}{l l|l}
        Context & Condition (c) & Template \\
        \hline
        \multirow{2}{*}{Pos} & ``Normal'' or ``Abnormal'' & The \{loclist\} is/are \{c\}. \\
        & Otherwise & There is \{c\} in the \{loclist\}. \\
        Neg & - & There is no \{c\}. \\
    \end{tabular}
    }\vspace{10pt} 
Here we show the 
Rules for creating synthetic sentences.
If there are multiple conditions in the sentence, we concatenate synthetic sentences for each of them.
The ``loclist'' is created by turning the list of anatomical locations associated with the condition/context into a natural language list (e.g., ``x,'' ``x and y,'' or ``x, y, and z''). We combine left- and right-side locations into one item (``left lung'' and ``right lung'' is mapped to ``lungs'').

\subsubsection{Synthetic Examples}

In Table \ref{tab:synthetic_sents_examples}, we present examples of synthetic examples formed via the rules in Section \ref{sec:synthetic_sents}.

\begin{table*}[]
    \centering
    \footnotesize
    \resizebox{\textwidth}{!}{
    \begin{tabular}{p{0.25\textwidth}p{0.12\textwidth}p{0.05\textwidth}p{0.25\textwidth}p{0.25\textwidth}}
        Original Sentence & Condition & Context & Location & Synthetic Sentence \\
        \hline
        Bulging mediastinum projecting over the left main bronchus and aortopulmonic window could be due to fat deposition exaggerated by low lung volumes. & low lung volumes & \greencheck & left lung, right lung & There is low lung volumes in the lungs. \\
        \hline
        \multirow{2}{=}{In the upper lobes, there is the suggestion of emphysema.} & abnormal & \greencheck & left mid lung zone, left upper lung zone, left lung, right mid lung zone, right upper lung zone & \multirow{2}{=}{The left lung, upper lung zones, and mid lung zones are abnormal. There is copd/emphysema in the lungs, upper lung zones, and mid lung zones.} \\
        \cline{2-4}
        & copd/emphysema & \greencheck & left mid lung zone, left upper lung zone, left lung, right mid lung zone, right upper lung zone & \\
        \hline
        Small left pleural effusion with atelectasis. & atelectasis & \greencheck & left costophrenic angle & There is atelectasis in the left costophrenic angle. \\
        \hline
        \multirow{2}{=}{No focal consolidation concerning for pneumonia.} & pneumonia & \xmark & left lung, right lung & \multirow{2}{=}{There is no pneumonia. There is no consolidation.} \\
        \cline{2-4}
        & consolidation & \xmark & right lung & \\
        \hline
        \multirow{3}{=}{Mild bibasilar atelectasis.} & abnormal & \greencheck & left lower lung zone, left lung, right lung, right lower lung zone & \multirow{3}{=}{The lungs and lower lung zones are abnormal. There is atelectasis in the lungs and lower lung zones. There is lung opacity in the lungs and lower lung zones.} \\
        \cline{2-4}
        & atelectasis & \greencheck & left lower lung zone, left lung, right lung, right lower lung zone & \\
        \cline{2-4}
        & lung opacity & \greencheck & left lower lung zone, left lung, right lung, right lower lung zone & \\
    \end{tabular}
    }
    \caption{Examples of \textbf{Synthetic Sentences}.}
    \label{tab:synthetic_sents_examples}
\end{table*}

\subsection{Metrics Figure}\label{sec:metric_demo}

\begin{figure}
    \centering
    \includegraphics[scale=.75]{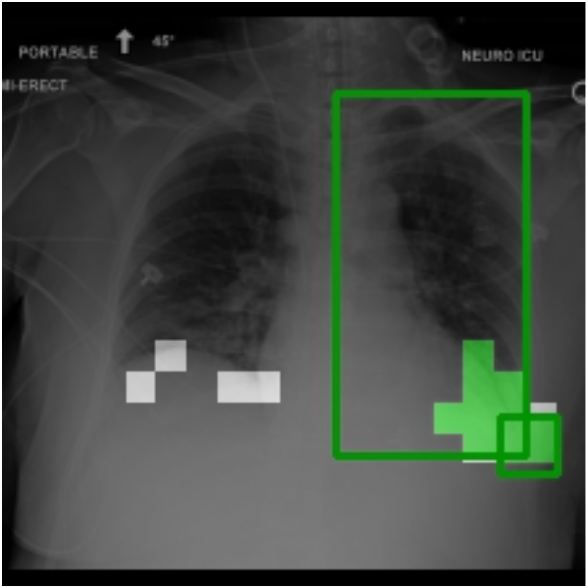}
    \caption{Visualization of Metrics. Attention map is thresholded, and {\color{green} true positives} are shown in green.}
    \label{fig:metric_demo}
\end{figure}

Figure \ref{fig:metric_demo} demonstrates what a thresholded (bilinearly upsampled) attention would look like and, for this specific threshold, which pixels are {\bf \color{green} true positives} (shown in green), false positives (shown in white), and false negatives (any other pixels inside either of the bounding boxes). For metrics such as \textbf{AUROC} and \textbf{Avg Precision}, statistics need to be computed while sliding through all possible thresholds.

\subsection{Perturbations Details}\label{sec:perturbations-details}

\begin{figure*}
    \centering
    \includegraphics[scale=.69]{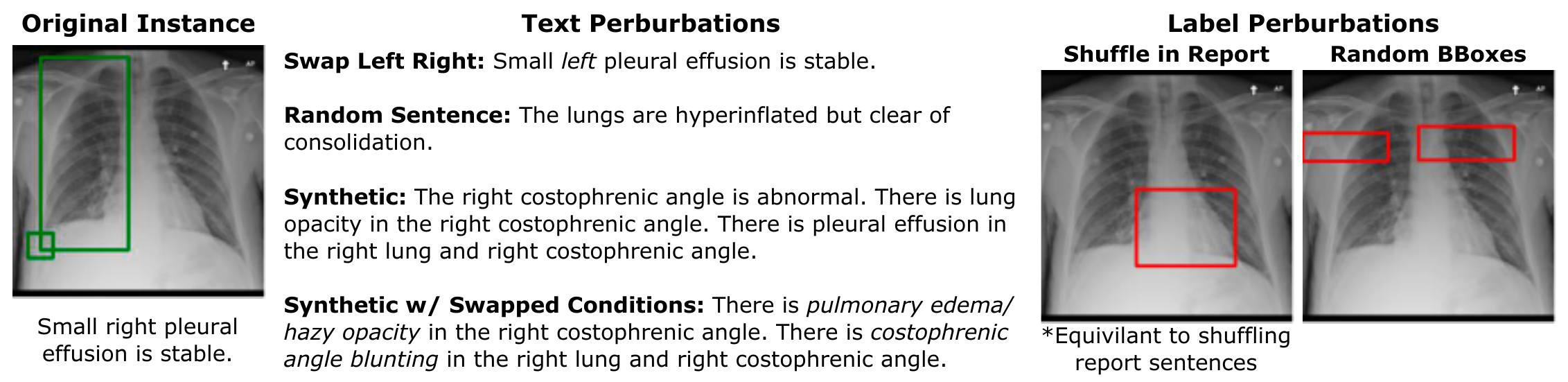}
    \caption{Examples of each \textbf{perturbation} (including \textbf{Snyth w/ Swapped Conditions}) for a given instance.}
    \label{fig:perturbations_app}
\end{figure*}

\vspace{0.5em}
\noindent \textbf{Swap Left Right} We replace every occurrence of the word ``right'' in the text with ``left'' and vice versa (ignoring capitalization). 
This 
is intended to probe 
the degree to which the attention mechanism relies on these two basic location cues.
Of course, many sentences 
do not contain these words because conditions (or lack thereof) occur on both sides of the chest X-ray. 
Therefore, it is particularly important to look at the metrics on the ``One Lung'' subset (Section \ref{sec:subsets}) for this perturbation.

\vspace{0.5em}
\noindent \textbf{Shuffle in Report} 
We shuffle the sets of bounding boxes for different sentences in the same report at random.
One would expect that performance would decrease significantly, because the resultant bounding boxes associated with given a sentence are (probably) wrong.
However, sentences within the same report \emph{might} be talking about similar regions.
Therefore, for this perturbation it is important to look at the instances where the overlap between (a) the region of interest for the sentence and (b) the regions associated with \emph{other} sentences in the report 
is low. 
We look at results for such cases explicitly using the {\bf Most Diverse Report BBoxes (MDRB)} subset (Section \ref{sec:subsets}).

\vspace{0.35em}
\noindent \textbf{Random Sentences} We replace sentences in an instance with other sentences, randomly drawn from the rest of the dataset.
Here too we expect performance to decrease significantly because the 
sampled text will
refer to an entirely different image.


\vspace{0.35em}
\noindent \textbf{Random BBoxes} We replace the set of bounding boxes for a sentence with a different set of bounding boxes randomly selected from the rest of the dataset.
This differs from the {\bf Random Sentences} perturbation in that the bounding boxes here are not only unrelated to the sentences, \emph{but also unrelated to the image}.
Therefore, we expect that this will have the poorest performance of all the settings, especially under the hypothesis that the attention is mostly a function of the image.

\vspace{0.35em}
\noindent \textbf{Synthetic+Swapped Conditions} This is performed on the synthetic, rather than original, sentences 
because swapping out conditions can only be done reliably when we generate sentences. 
To swap conditions, 
we follow the same rules for generating the synthetic sentence with a different condition randomly sampled from a set of (other) possible conditions.
Possible conditions are defined as any condition (excluding the current) that occurs in the same anatomical locations anywhere else in the gold dataset.\footnote{If there are no other conditions, we leave the condition as is and the synthetic sentence is not perturbed.}
This perturbation should measure the impact of conditions on model attention. 

\subsection{Subset Details}\label{sec:subset_details}

\vspace{0.35em}
\noindent \textbf{Abnormal} Image/sentence pairs where there is an ``abnormal'' label associated with the sentence. This occurs if any conditions are mentioned in a \emph{positive} context, i.e., where the radiologist believes the patient has said condition.
This targets ``interesting'' examples where the attention 
should ideally highlight the region relevant to the condition described. 

\vspace{0.35em}
\noindent \textbf{One Lung} Image/sentence pairs where the bounding boxes corresponding to the sentence contain a bounding box of either the left or right lung, but not both.
This subset allows us to evaluate how the model performs when the attention should only be on one side of the image.

\vspace{0.35em}
\noindent \textbf{Most Diverse Report BBoxes} Instances where the overlap in the sets of bounding boxes for sentences within the same report is minimal. 
Specifically, we calculate the mean intersection over union (IOU; Section \ref{sec:data_and_metrics}) of the segmentation labels $\ell_1, \ell_2$ for pairs of sentences in the same report.
We then take the 10\% of instances within reports with the smallest mean IOU.
This subset is intended to include examples within reports where multiple distinct regions of interest discussed in different sentences.

These first two subsets are important because in many examples there is nothing abnormal, and the reports contain sentences such as ``No effusion is present.''
For these types of sentences, the bounding boxes are commonly over both lungs because the evidence for the sentence is that nothing abnormal is in either lung.
In these situations, it seems as though it might be easier for the model to realize higher scores for two reasons: 1) lungs take up most of the image, so attention is likely to fall in the bounding boxes, and 2) the lungs are a pretty good guess for the ``important'' regions of any image, independent of the text.
The last subset is important because it comprises examples which contain a set of target bounding boxes and associated texts which cover mostly distinct image regions.

\subsection{Annotations}\label{sec:annotations_appendix}

\begin{figure*}
    \centering
    \includegraphics[scale=.31]{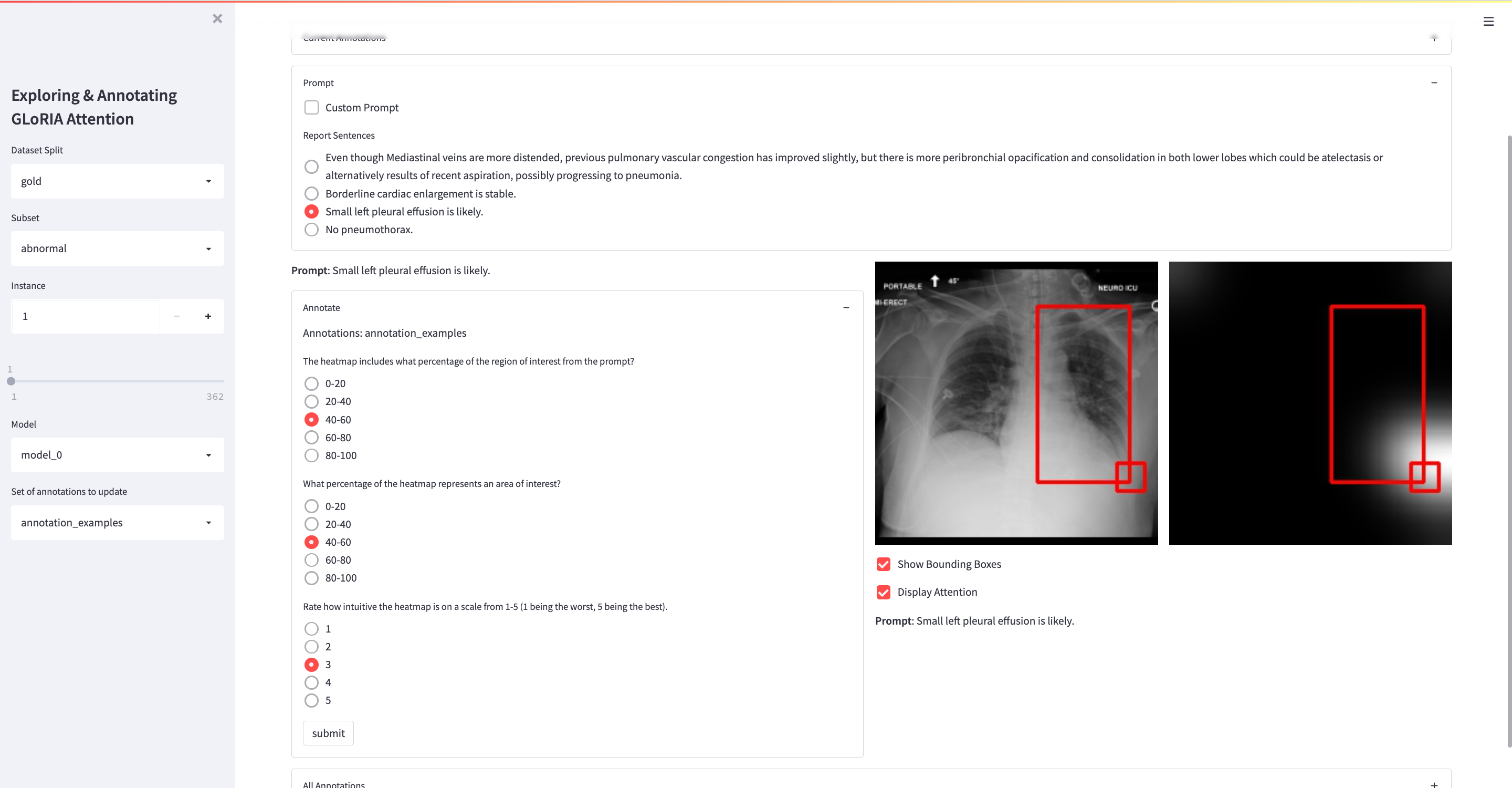}
    \caption{Interface}
    \label{fig:interface}
\end{figure*}

In Figure \ref{fig:interface}, we present our user interface for collecting annotations created using streamlit.
In Section \ref{sec:instructions} (below), we show the annotation instructions.

\subsubsection{Instructions}\label{sec:instructions}

Our aim here is to collect judgements (*annotations*) concerning the interpretability and possible usefulness of alignments between text snippets and image regions induced by neural network models.
More specifically, we will ask you to evaluate “heatmaps” output by different unsupervised (or minimally supervised) models which attempt to align natural language (sentences) and image regions (within accompanying chest X-rays).
We ask three specific questions to asses these heatmaps; each question is 5-way multiple choice, and each of the answers are described below.
In each round of annotation collection, we aim to collect annotations for multiple models with respect to a shared set of text snippets.
That is, for each image, we ask for multiple assessments (across models) for the quality of alignments performed for a particular sentence.
You will not be told which model generated which "heatmaps", and model aliases are randomly selected for every instance.

\subsubsection*{Prompts}

You can choose the natural language sentences fed to the model—which we refer to as “prompts”— by either selecting a sentence from the list of sentences in the associated radiology report, or by writing your own ``custom'' prompt.
We ask you to complete one round of annotations for report sentences, followed by one round in which you evaluate the alignments generated by the model for custom prompts (i.e., text you enter).
For the report sentences round, we ask you to select one sentence that you think is interesting from the list of report sentences (prior to looking at any heatmaps).
More specifically, you should, when possible, select a sentence with a focal abnormality that has strong clinical relevance.
If one is not present, you can select a sentence that has a more diffuse abnormality or a negative statement that is still relatively focal.
You will then annotate or judge the alignments induced by all models for this particular sentence.
For instances that you do not think have any appropriate sentences or for instances where you can think of a better prompt, we ask you to write a prompt to annotate using the ``custom prompt'' option in addition to annotating the best sentence from the report.

\subsubsection*{Annotations}

Bellow we list the questions and what each of the possible answers would mean.
\begin{enumerate}
    \item \textbf{The heatmap includes what percentage of the region of interest from the prompt?}
    \begin{itemize}
        \item 0-20 -- The heatmap is focused on entirely the wrong part of the image, does not highlight any part of the image strongly, or has very minimal intensity on the region of interest.
        \item 20-40
        \item 40-60 -- The heatmap comes close to covering the region of interest, or does cover the region of interest but with not too much intensity.
        \item 60-80
        \item 80-100 -- The prompt refers to a region that is within a high-intensity part of the heatmap.
    \end{itemize}
    \item \textbf{What percentage of the heatmap represents an area of interest?}
    \begin{itemize}
        \item 0-20 -- This heatmap is all over the place or highlights a large portion of the image.
        \item 20-40
        \item 40-60 -- The focus includes the relevant region(s) but also other irrelevant regions (either adjacent or elsewhere in the image).
        \item 60-80
        \item 80-100 -- The heatmap is very targeted to only the parts of the image most relevant to the prompt.
    \end{itemize}
    \item \textbf{Rate how intuitive the heatmap is on a scale from 1-5 (1 being the worst, 5 being the best).}
    \begin{itemize}
        \item 1 -- The heatmap is completely unhelpful, counterintuitive, or misleading.
        \item 2 -- The heatmap might have something in common with an intuitive one, but very little.
        \item 3 -- The heatmap does show a region of tiniest, but has some stray parts or does not catch all relevant regions.
        \item 4 -- The heatmap is reasonably intuitive and contains mostly (though not exclusively) the regions I would expect.
        \item 5 -- The heatmap is almost exactly what you might draw to represent the region of interest.
    \end{itemize}
\end{enumerate}

\subsubsection*{Ground truth bounding boxes}

You have the ability to see \textit{ground truth} bounding boxes from the dataset associated with the particular sentence you have selected from the report; these were manually drawn to match the corresponding sentence.
We suggest that you use these bounding boxes when annotating the heatmaps associated with the report sentences.
No such bounding boxes are available for the custom prompts that you will author.


\subsection{No Attn Model Saturating Attn Map}\label{sec:no_attn_sat}

\begin{figure}
    \centering
    \includegraphics[scale=.75]{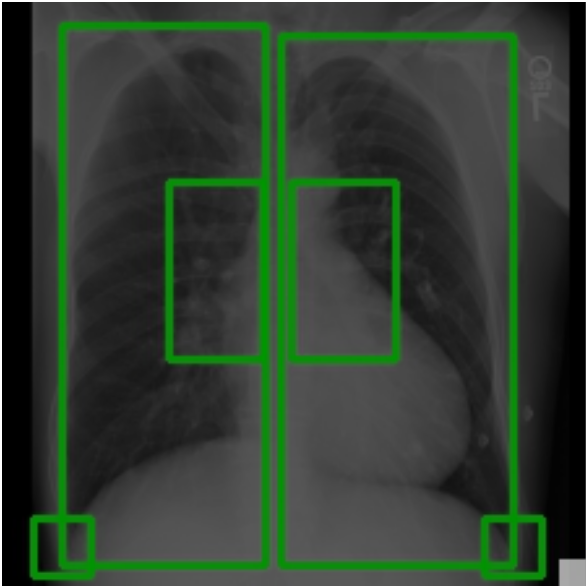}
    \caption{Example of when the attention map produced by the {\bf + ``No Attn'' Token} model  is fully saturated by having strong attention on the ``No Attn'' token. The bottom right corner depicts the strength of the attention on the ``No Attn'' token, and the rest of the attention map is invisible because it has little attention in comparison.}
    \label{fig:no_attn_sat}
\end{figure}

Figure \ref{fig:no_attn_sat} depicts what happens when the model attends very highly to the ``No Attn'' token.

\section{Full/Additional Results}\label{sec:results_app}

Here we include full/expanded results for the tables in the main paper and some additional results from which we may not yet have a takeaway.

\subsection{Dropping Large Bounding Boxes for Evaluation}\label{sec:dropping_large_bboxes}

As discussed in sections \ref{sec:alignment_accuracy} and \ref{sec:localization}, we noticed that in many cases in the ImaGenome dataset, the bounding boxes cover more than the true region of interest.
We argue the original labels still serve us well in understanding when highlighted regions are far from where they should be, but to get a better sense of the precision of the models, we also trim out some of the larger boxes and repeat the evaluation from Tables \ref{tab:gloria_localization_performance} and \ref{tab:retrained_localization_performance} with the modified labels in Table \ref{tab:retrained_localization_performance_trimmed_labels}.

\begin{table}[]
    \centering
    \begin{tabular}{l l l}
    \hline
        Model & AUROC & Avg. P \\
    \hline
GLoRIA & \textbf{64.67} & \textbf{32.79} \\
GLoRIA Retrained & 54.56 & 26.00 \\
\hline
+Word Masking & 59.88 & 28.62 \\
+Clinical Masking & 53.75 & 25.42 \\
+"No Attn" Token & 57.10 & 28.43 \\
\hline
+Abnormal & 57.83 & 29.42 \\
+30-shot Finetuned & \textbf{61.98} & \textbf{35.27} \\
    \hline
    \end{tabular}
    \caption{Localization performance with large bounding boxes trimmed.} 
    \label{tab:retrained_localization_performance_trimmed_labels}
\end{table}

Specifically for a sentence's label, we delete the ``right lung''/``left lung'' bounding box when there exist another bounding box within the label that contains the word ``right''/``left''.
If no other box exists on the same side, then we still keep the full lung bounding box.
As an example, in Figure \ref{fig:qualitative} the larger of the two bounding boxes would be deleted from the label.

\subsection{Custom Prompts}

\begin{figure*}
    \centering
    \includegraphics[scale=.6]{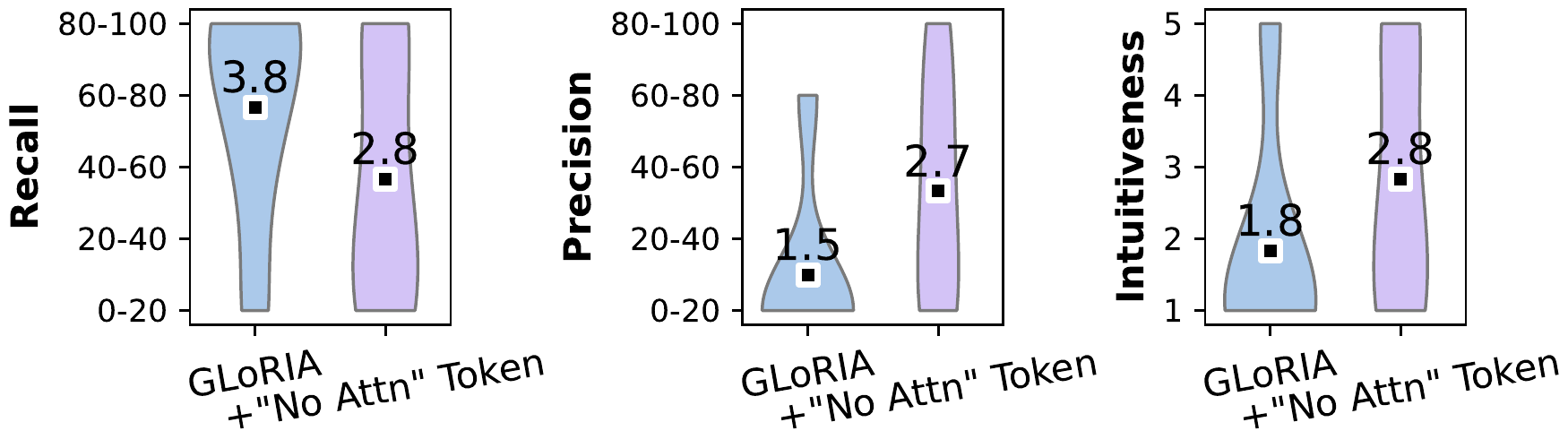}
    \caption{Custom annotation results (means over 6 instances).}
    \label{fig:custom_anns}
\end{figure*}

We also let annotators chose to write (and annotate) a fitting prompt if one was not present in the report.
Figure \ref{fig:custom_anns} shows the annotations for these ``custom'' prompts for \textbf{GLoRIA} and the \textbf{+``No Attn'' Token} models and indicates that even in this small, potentially out of domain setting, the scores are consistent with the in-domain annotations.

\subsection{Localization performance for all models on all subsets.}\label{sec:localization_app}

Table \ref{tab:retrained_localization_performance_subsets} reports additional results to those in Table \ref{tab:retrained_localization_performance}, describing localization performance on each subset individually.

Not shown in the main paper, we can see here that synthetic sentences perform comparably to real sentences, validating our method for constructing synthetic sentences.
In fact, on \textbf{+30-shot Finetuned}, there is a significant jump in performance when using synthetic sentences.

\begin{table*}[]
    \centering
    \footnotesize
    \begin{tabular}{l c l l l l l l l l}
    \hline
        \multirow{2}{*}{Model} & \multirow{2}{*}{Synth} & \multicolumn{2}{c}{All} & \multicolumn{2}{c}{Abnormal} & \multicolumn{2}{c}{One Lung} & \multicolumn{2}{c}{MDRB} \\
        & & AUROC & Avg. P & AUROC & Avg. P & AUROC & Avg. P & AUROC & Avg. P \\
        \hline
\multirow{2}{*}{UNITER*} & \xmark & 84.92 & 68.57 & 83.47 & 66.33 & 76.86 & 57.71 & 80.49 & 56.12 \\
 & \greencheck & 84.87 & 68.80 & 83.68 & 67.10 & 76.61 & 57.64 & 79.96 & 56.11 \\
\hline
\multirow{2}{*}{GLoRIA} & \xmark & 69.07 & 51.68 & 69.51 & 48.29 & 65.48 & 38.68 & 65.01 & 36.96 \\
 & \greencheck & 69.28 & 52.17 & 70.30 & 49.93 & 66.62 & 41.29 & 66.24 & 37.95 \\
\hline
\multirow{2}{*}{GLoRIA Retrained} & \xmark & 55.84 & 41.22 & 55.11 & 37.01 & 53.45 & 28.87 & 55.14 & 30.36 \\
 & \greencheck & 54.98 & 41.05 & 53.39 & 36.59 & 52.59 & 28.67 & 54.95 & 30.22 \\
\hline
\multirow{2}{*}{+Word Masking} & \xmark & 61.44 & 44.69 & 61.80 & 41.42 & 58.14 & 31.95 & 60.23 & 32.54 \\
 & \greencheck & 59.28 & 43.36 & 58.47 & 39.32 & 56.00 & 30.62 & 57.61 & 30.87 \\
\hline
\multirow{2}{*}{+Clinical Masking} & \xmark & 54.67 & 40.61 & 54.94 & 37.30 & 52.78 & 28.73 & 54.27 & 29.20 \\
 & \greencheck & 54.57 & 40.60 & 53.62 & 36.52 & 51.70 & 28.22 & 53.91 & 28.99 \\
\hline
\multirow{2}{*}{+"No Attn" Token} & \xmark & 57.00 & 41.80 & 57.32 & 39.20 & 56.47 & 32.65 & 56.76 & 31.08 \\
 & \greencheck & 56.29 & 41.66 & 56.62 & 38.92 & 56.09 & 32.69 & 56.18 & 30.84 \\
\hline
\multirow{2}{*}{+Abnormal} & \xmark & 55.89 & 43.42 & 57.59 & 42.20 & 54.68 & 33.01 & 55.33 & 32.32 \\
 & \greencheck & 52.78 & 41.86 & 54.05 & 39.96 & 51.22 & 30.48 & 53.15 & 31.01 \\
\hline
\multirow{2}{*}{+30-shot Finetuned} & \xmark & 63.90 & 52.80 & 65.28 & 50.44 & 61.61 & 40.79 & 62.16 & 39.91 \\
 & \greencheck & 68.38 & 56.05 & 73.14 & 55.57 & 67.92 & 45.00 & 66.26 & 42.28 \\
\hline
\multirow{2}{*}{+Rand Sents} & \xmark & 38.88 & 30.55 & 41.10 & 28.16 & 41.15 & 22.45 & 41.47 & 21.60 \\
 & \greencheck & 36.09 & 29.15 & 39.84 & 27.73 & 36.81 & 20.76 & 39.73 & 20.77 \\
        \hline
    \end{tabular}
    \caption{\textbf{Localization performance} for each retrained model on the subsets. This also includes results on synthetic sentences (Section \ref{sec:synthetic_sents}).}
    \label{tab:retrained_localization_performance_subsets}
\end{table*}

\subsection{Deltas of all models on all subsets}\label{sec:deltas_app}

\begin{figure*}
    \centering
    \includegraphics[scale=.55]{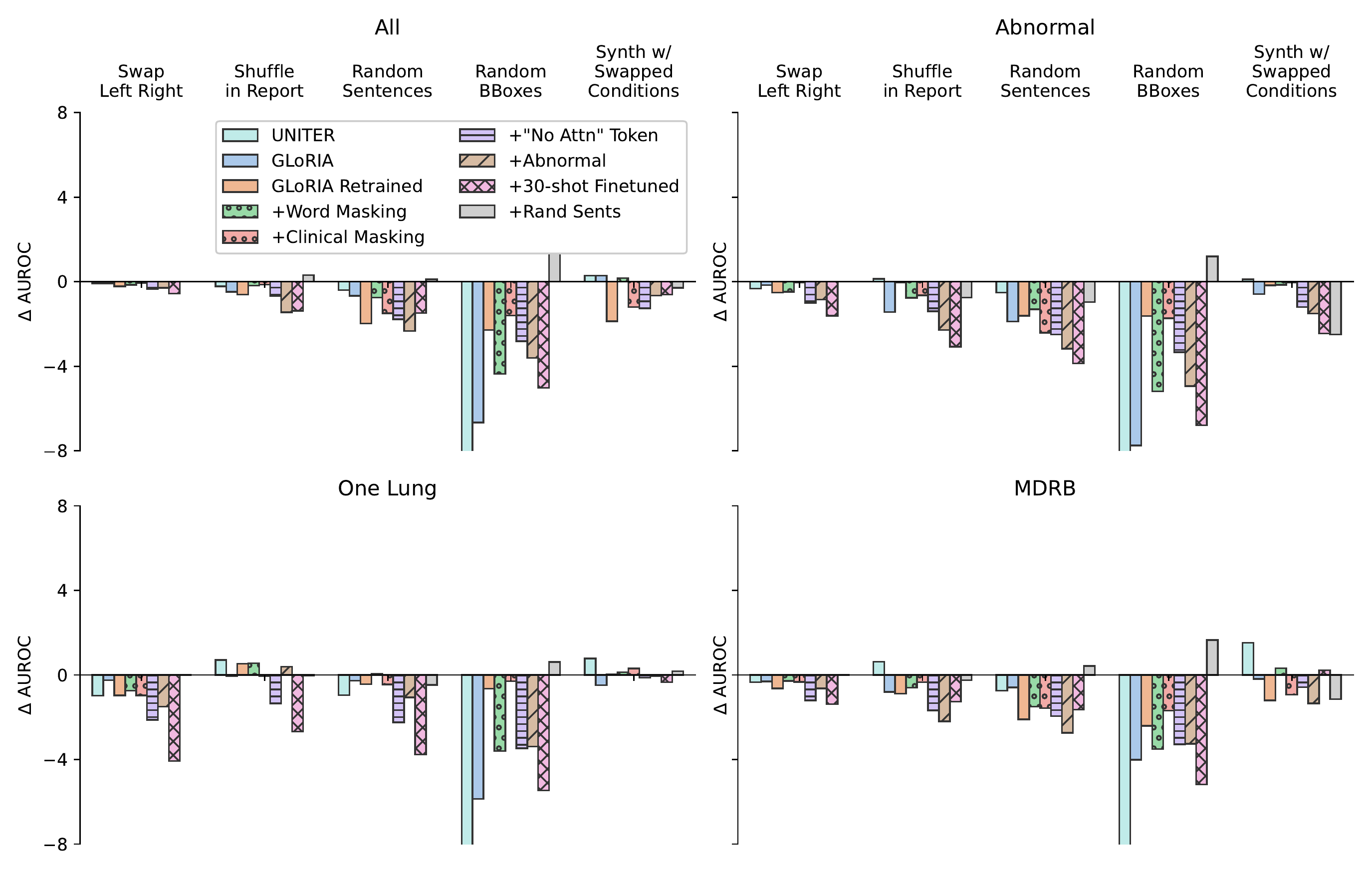}
    \caption{$\Delta$\textbf{AUROC} for all models and subsets.}
    \label{fig:deltas_retrained_full}
\end{figure*}

In Figure \ref{fig:deltas_retrained_full} we report results analogous to those in Figures \ref{fig:deltas_gloria} and \ref{fig:deltas_retrained}, but on all subsets, all models, and all perturbations at once.

The results from swapping conditions in synthetic sentences, which were not shown in the main paper, 
vary across 
data subsets (Figure \ref{fig:deltas_retrained_full}).
The most telling subset for this perturbation is probably the \textbf{Abnormal} set.
The results here are 
difficult to interpret because the \textbf{+Rand Sents} model seems to be considerably effected, which is counter-intuitive as we would expect this model to be invariant to the text by construction (note that the other perturbation results are consistent with this).
Given this, we do not draw any particular conclusions from the swapped conditions experiment at present.

\subsection{$\Delta$ Average Precision}

\begin{figure*}
    \centering
    \includegraphics[scale=.55]{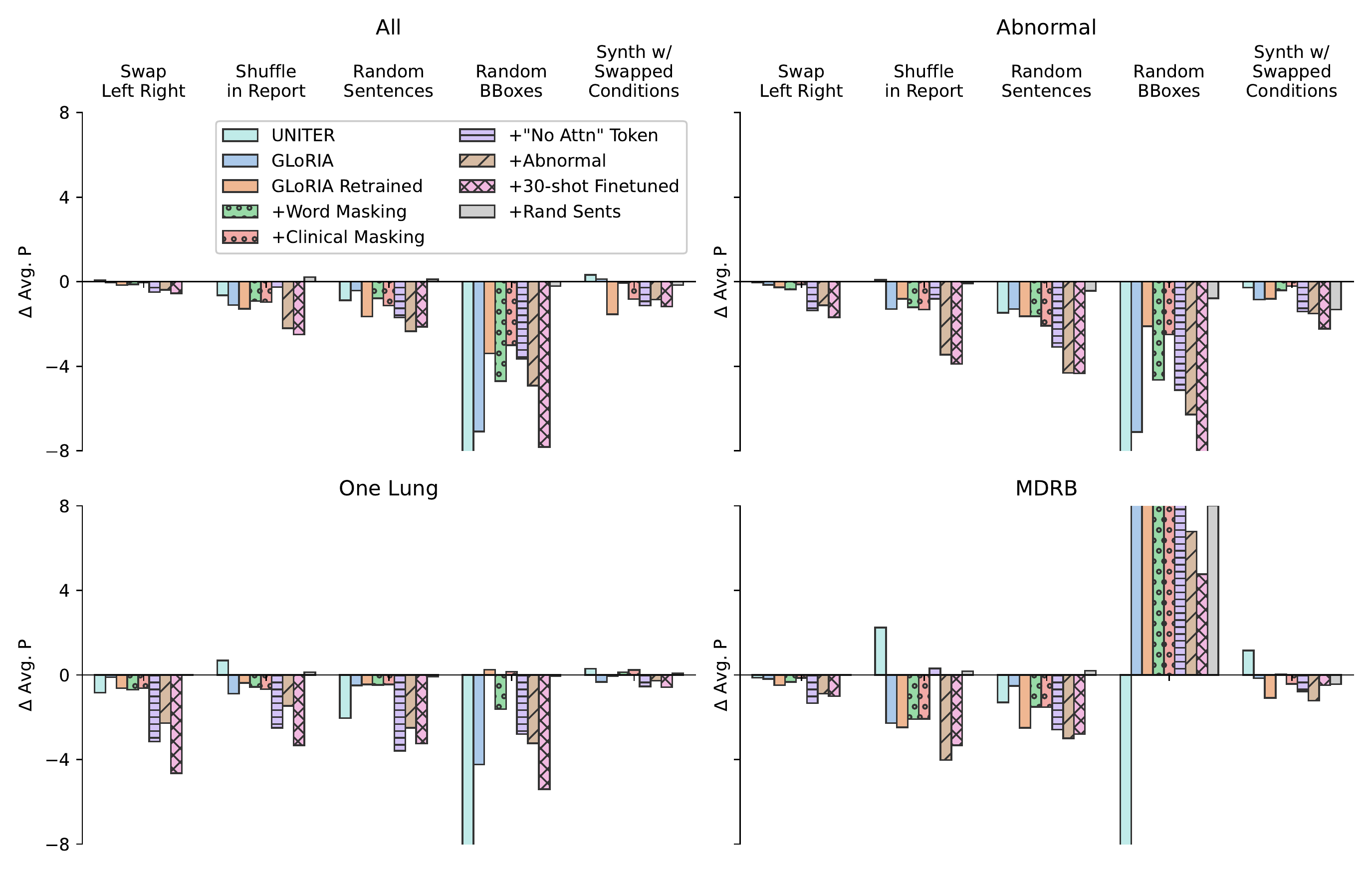}
    \caption{\textbf{$\Delta$Average Precision} for all models and subsets.}
    \label{fig:deltas_retrained_avgp}
\end{figure*}

In Figure \ref{fig:deltas_retrained_avgp}, we plot the analogous plot to Figure \ref{fig:deltas_retrained_full} for the changes in Average Precision as opposed to AUROC.
Average Precision seems to tell a similar story to AUROC in terms of which models have greater changes for each perturbation.
The only major difference is that for Average Precision, all models show a positive change for the Random BBoxes perturbation in the MDRB subset.
This is likely because picking a random bounding box from the whole dataset when in this subset means that the random bounding box will likely be bigger than the original because the bounding boxes in this subset tend to be small.
Having a larger bounding box as a label would therefore likely improve precision in general.
This makes it harder to interpret this particular perturbation in this subset.

\subsection{Random Attention KL Divergences}

To measure the extent to which a model eschews the text and relies mostly on the image to induce an attention pattern, 
we introduce  
\textbf{Random Attention KL Divergence}. 
This is the symmetric Kullback–Leibler (KL) divergence for an instance between (a) the attention distribution induced given the original text, and (b) the attention over the same image but paired with random text.
In Table \ref{tab:rand_attn_kl_div_subsets}, we show the mean \textbf{Random Attention KL Divergence} for each subset.

\begin{table*}[]
    \footnotesize
    \centering
    \begin{tabular}{l l l l l}
    \hline
Model & All & Abnormal & One Lung & MDRB \\
\hline
UNITER & 0.04 & 0.04 & 0.04 & 0.04 \\
GLoRIA & 0.08 & 0.08 & 0.08 & 0.09 \\
Retrained & 0.04 & 0.04 & 0.03 & 0.04 \\
+Word Masking & 0.05 & 0.04 & 0.04 & 0.05 \\
+Clinical Masking & 0.03 & 0.03 & 0.02 & 0.03 \\
+``No Attn'' Token & 0.04 & 0.05 & 0.04 & 0.04 \\
+Abnormal & 0.11 & 0.11 & 0.10 & 0.11 \\
+30-shot Finetuned & \textbf{0.17} & \textbf{0.16} & \textbf{0.15} & \textbf{0.17} \\
+Rand Sents & 0.00 & 0.00 & 0.00 & 0.00 \\
\hline
    \end{tabular}
    \caption{Average \textbf{Random Attention KL Divergences} on the subsets}
    \label{tab:rand_attn_kl_div_subsets}
\end{table*}

\subsection{Candidate Selection Accuracy for other subsets}

In \ref{tab:candidate_selection_subsets}, we extend Table \ref{tab:candidate_selection} to the remaining subsets.

\begin{table*}[]
    \footnotesize
    \centering
    \begin{tabular}{l l l l l}
    \hline
        Model & \multicolumn{2}{c}{{\bf One Lung}} & \multicolumn{2}{c}{{\bf MDRB}} \\
        & local & global & local & global \\
        \hline
UNITER & - & 70.1 & - & 65.5 \\
GLoRIA & 38.9 & 72.3 & 53.6 & 73.8 \\
GLoRIA Retrained & 62.8 & 86.7 & 75.4 & \textbf{84.1} \\
+Word Masking & \textbf{82.5} & \textbf{88.4} & \textbf{78.6} & 81.7 \\
+Clinical Masking & 60.0 & 83.2 & 67.9 & 82.5 \\
+``No Attn'' Token & 70.9 & 83.9 & 69.0 & 80.2 \\
+Abnormal & 72.3 & 85.6 & 73.0 & 75.4 \\
+30-shot Finetuned & 59.6 & 84.9 & 65.9 & 79.0 \\
+Rand Sents & 44.6 & 59.6 & 50.8 & 48.4 \\
\hline
    \end{tabular}
    \caption{\textbf{Candidate Selection Accuracy} for other subsets.} 
    \label{tab:candidate_selection_subsets}
\end{table*}

\subsection{Entropy}

In Table \ref{tab:attn_entropy} we present results for the entropy attention mechanisms for each model for the entire dataset as well as the subsets.

\begin{table*}[]
    \footnotesize
    \centering
    \begin{tabular}{l r r r r}
Model & \textbf{All} & \textbf{Abnormal} & \textbf{One Lung} & \textbf{MDRB} \\
\hline
UNITER* & \textbf{1.777} & \textbf{1.668} & \textbf{1.644} & \textbf{1.721}\\
GLoRIA & 5.828 & 5.841 & 5.833 & 5.822\\
GLoRIA Retrained & 5.857 & 5.863 & 5.872 & 5.862\\
+Word Masking & 5.841 & 5.848 & 5.858 & 5.846\\
+Clinical Masking & 5.864 & 5.866 & 5.876 & 5.868\\
+``No Attn'' Token & 5.849 & 5.855 & 5.861 & 5.856\\
+Abnormal & 5.803 & 5.816 & 5.825 & 5.806\\
+30-shot Finetuned & \textbf{5.677} & \textbf{5.729} & \textbf{5.748} & \textbf{5.692} \\
+Rand Sents & 5.889 & 5.889 & 5.889 & 5.889\\
    \end{tabular}
    \caption{\textbf{Attention Entropy}}
    \label{tab:attn_entropy}
\end{table*}

\subsection{Performance across Specific Abnormalities}

\begin{figure}
    \centering
    \includegraphics[scale=.43]{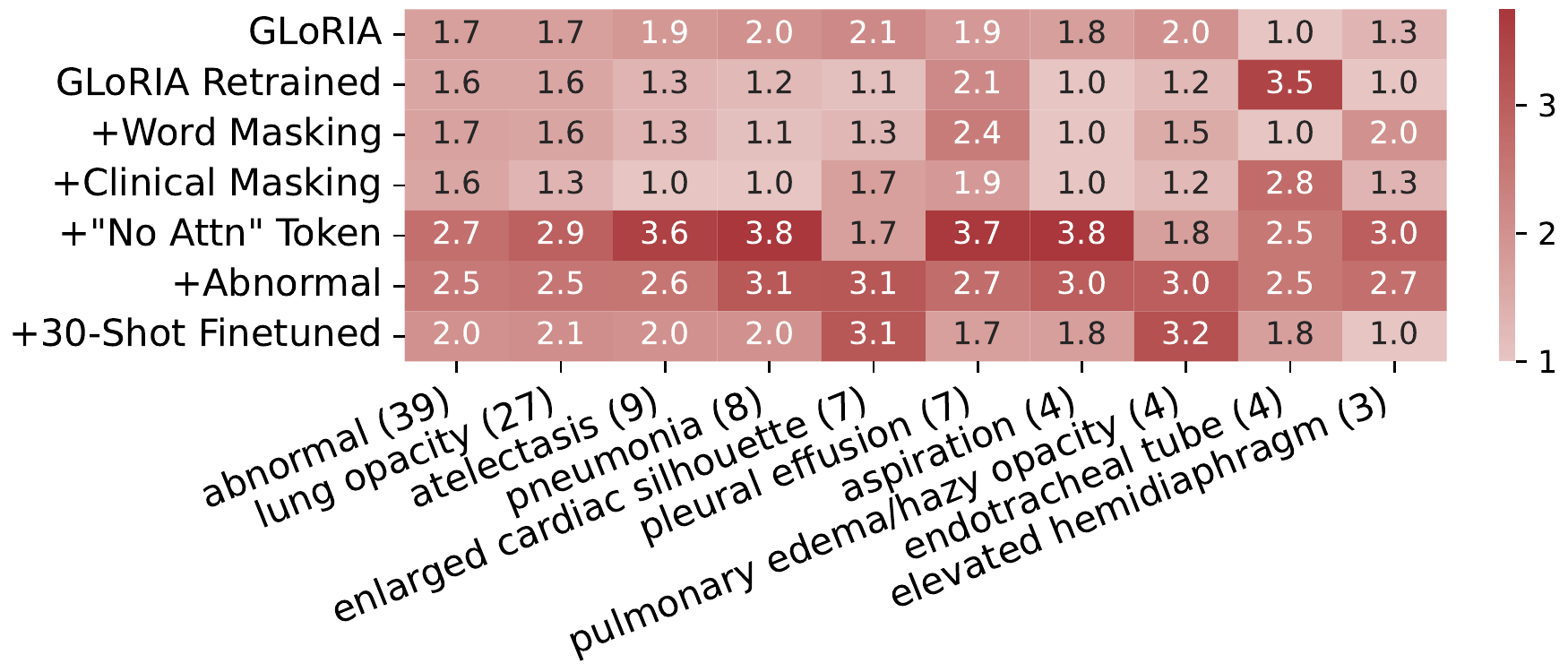}
    \caption{\textbf{Intuitiveness} on subsets of the annotations corresponding to the top 10 most frequent abnormalities.}
    \label{fig:intuitiveness_on_subsets}
\end{figure}

In Figure \ref{fig:intuitiveness_on_subsets}, we present Intuitiveness for all models on examples with specific abnormalities.

\subsection{Correlations}

\begin{figure}
    \centering
    \includegraphics[scale=.6]{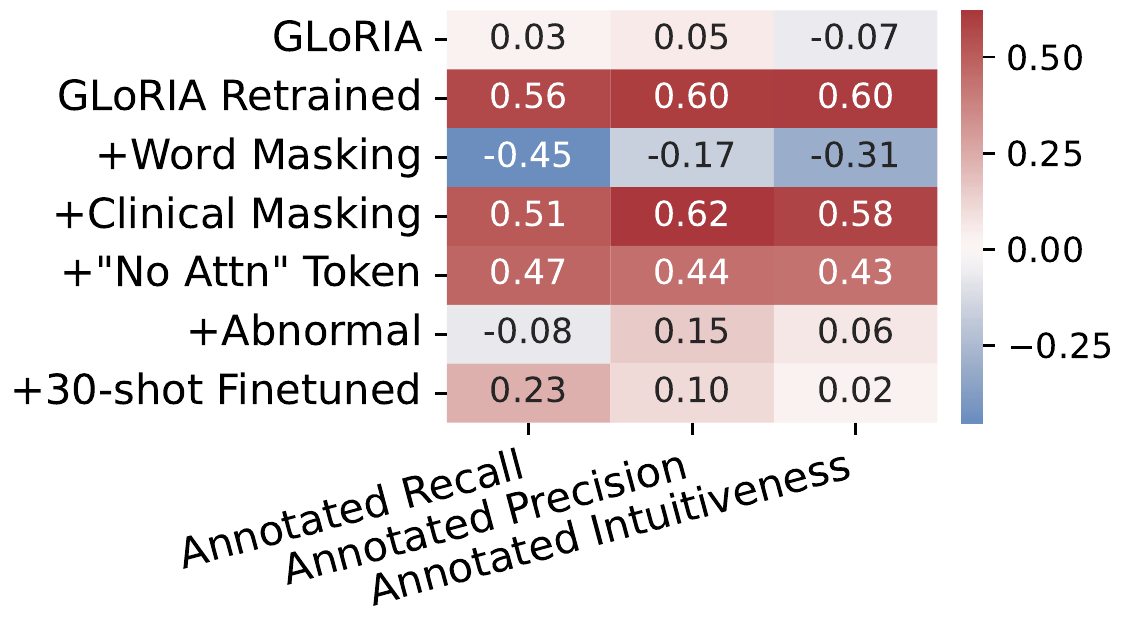}
    \caption{\textbf{Correlations with local similarity} (from heatmaps below).}
    \label{fig:correlations_localsim}
\end{figure}
\begin{figure*}
    \centering
    \includegraphics[scale=.48]{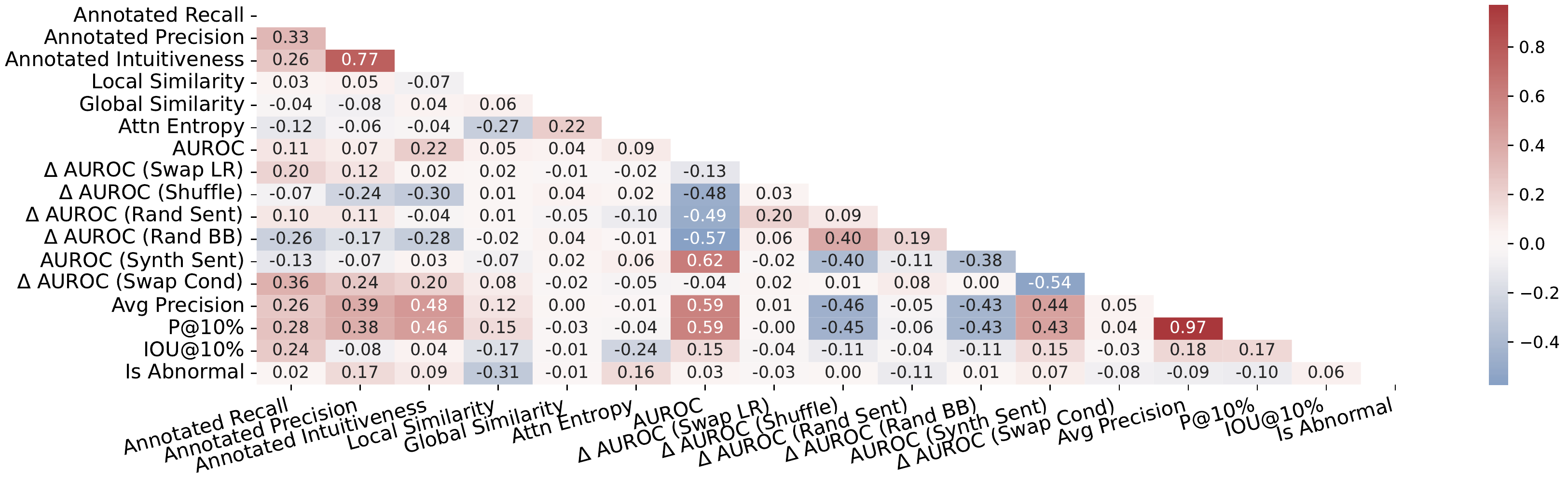}
    \caption{GLoRIA}
    \label{fig:corr_orig}
\end{figure*}
\begin{figure*}
    \centering
    \includegraphics[scale=.48]{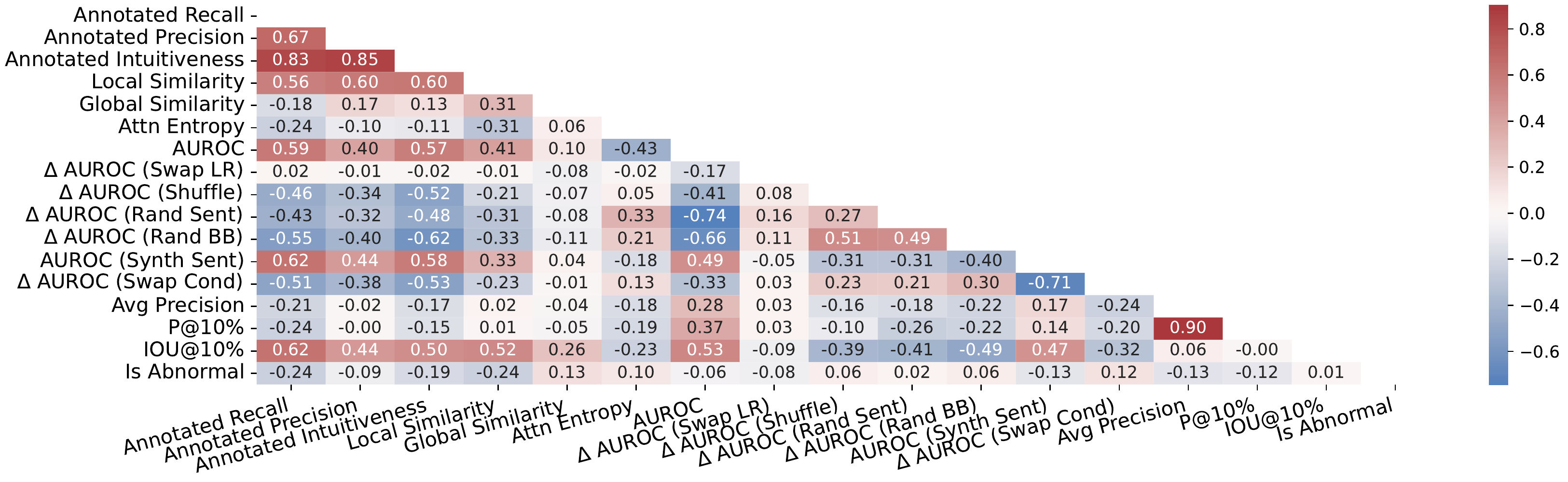}
    \caption{GLoRIA Retrained}
    \label{fig:corr_retr}
\end{figure*}
\begin{figure*}
    \centering
    \includegraphics[scale=.48]{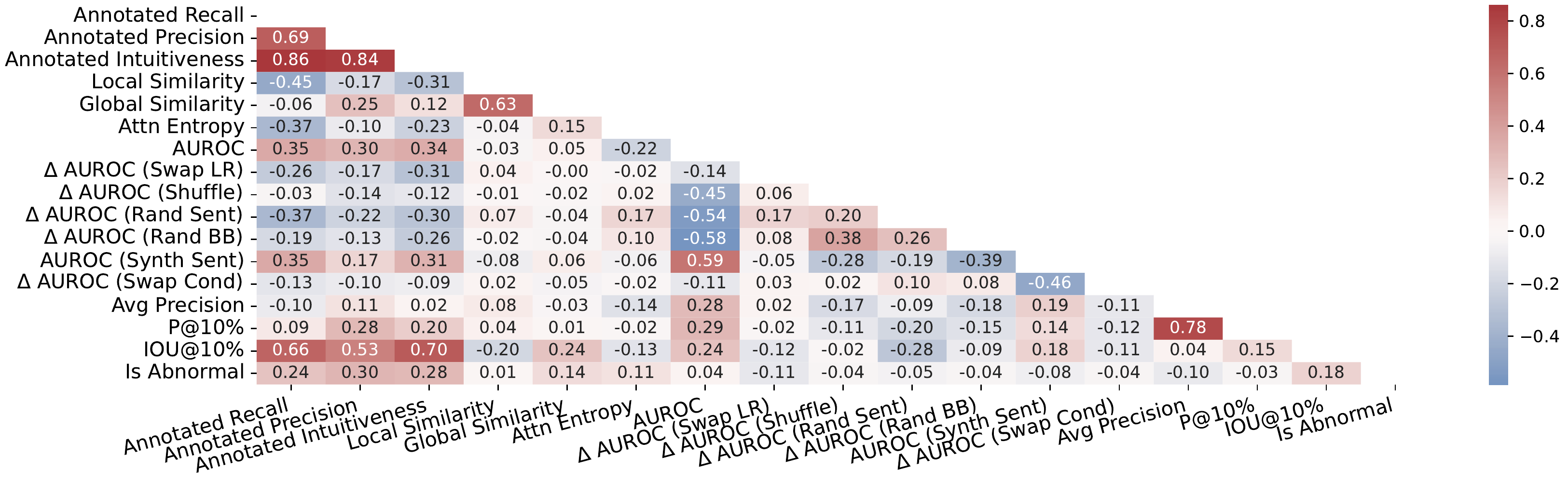}
    \caption{+Word Masking}
    \label{fig:corr_wm}
\end{figure*}
\begin{figure*}
    \centering
    \includegraphics[scale=.48]{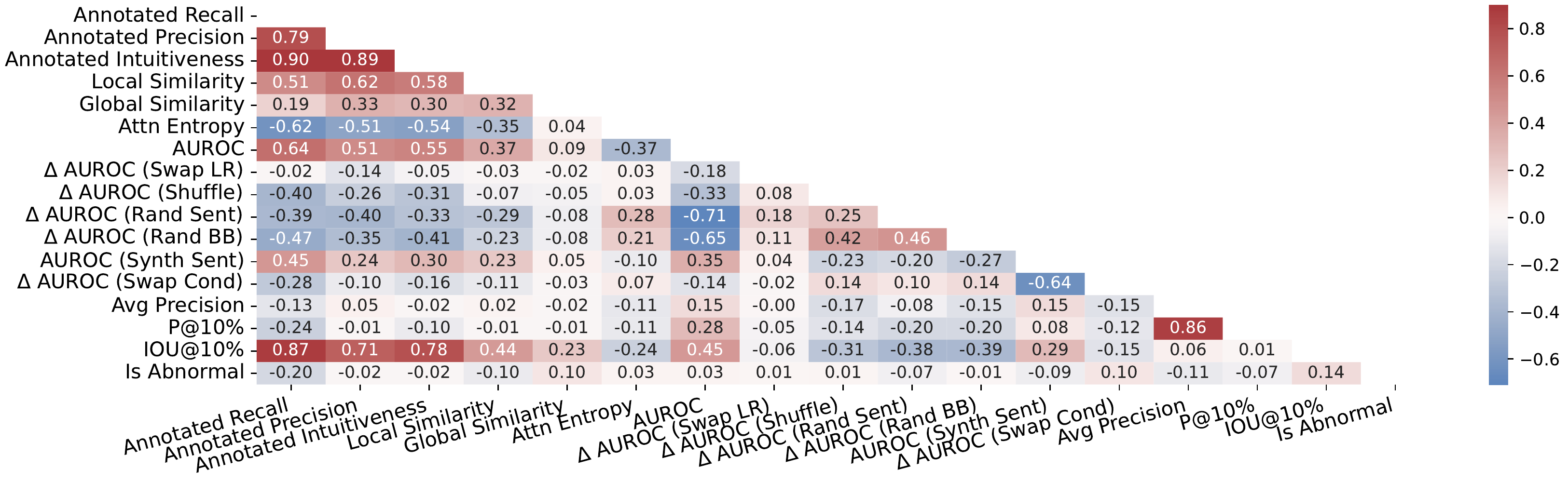}
    \caption{+Clinical Masking}
    \label{fig:corr_cm}
\end{figure*}
\begin{figure*}
    \centering
    \includegraphics[scale=.48]{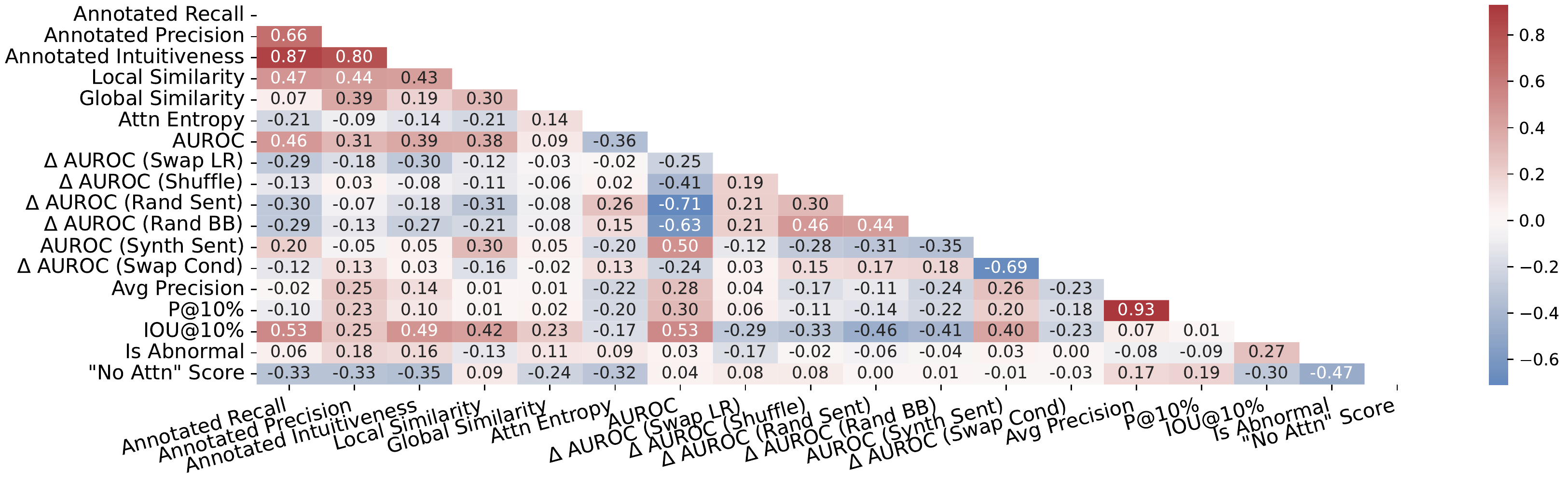}
    \caption{+``No Attn'' Token}
    \label{fig:corr_na}
\end{figure*}
\begin{figure*}
    \centering
    \includegraphics[scale=.48]{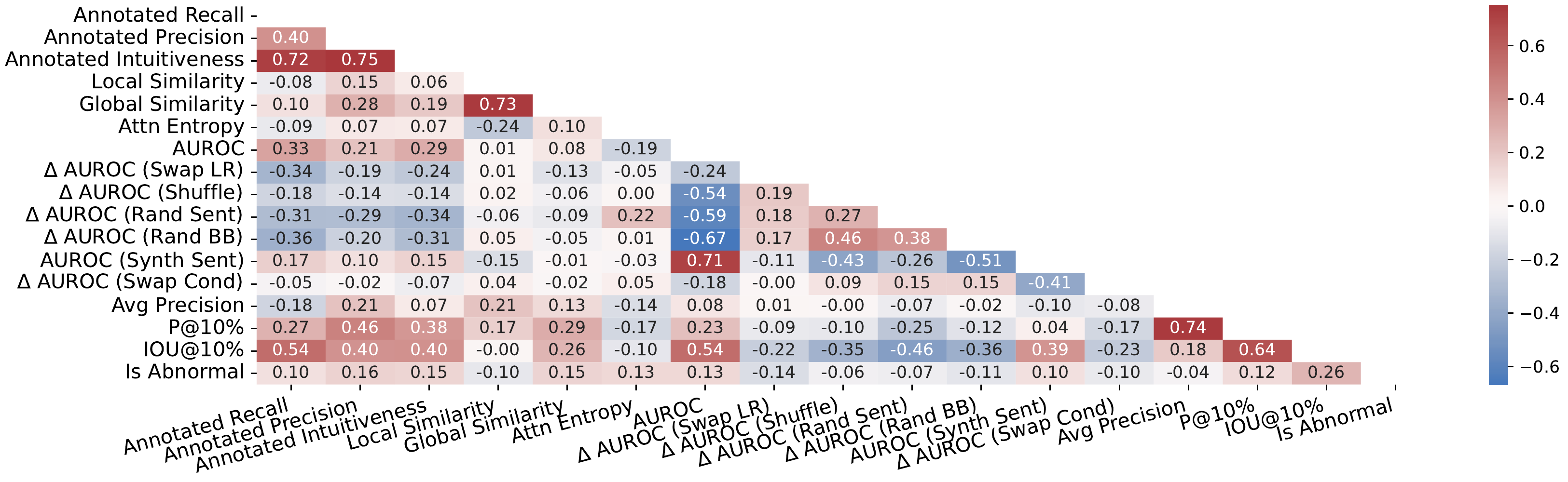}
    \caption{+Abnormal}
    \label{fig:corr_ab}
\end{figure*}
\begin{figure*}
    \centering
    \includegraphics[scale=.48]{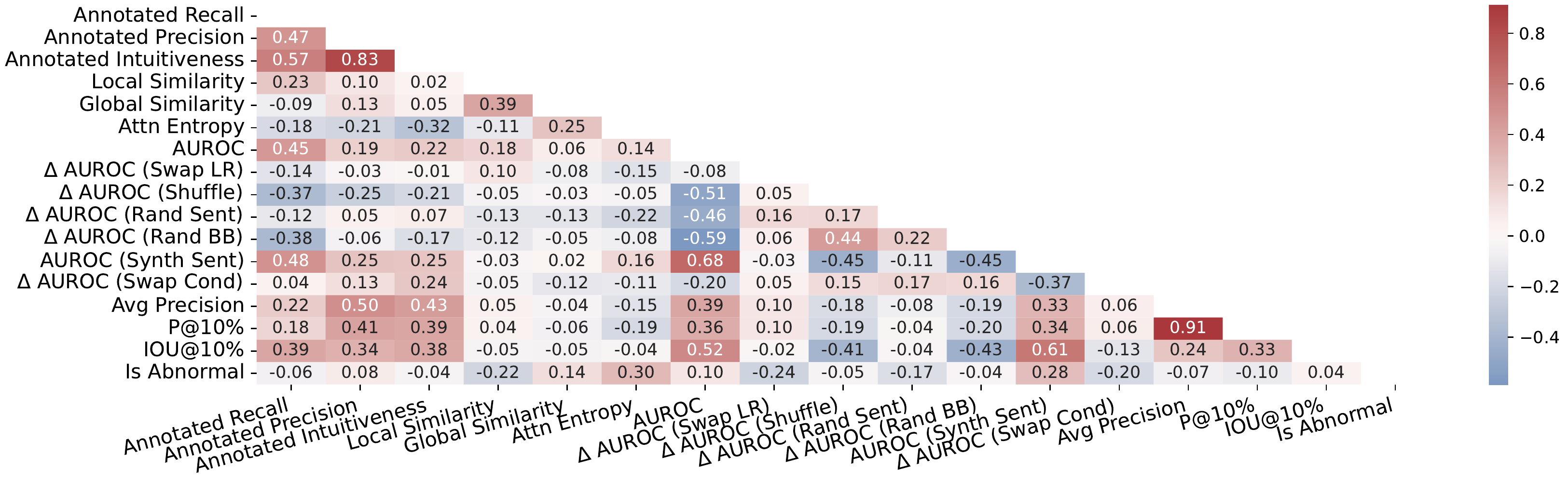}
    \caption{+30-shot Finetuned}
    \label{fig:corr_fi}
\end{figure*}

In Figures \ref{fig:corr_orig}, \ref{fig:corr_retr}, \ref{fig:corr_wm}, \ref{fig:corr_cm}, \ref{fig:corr_na}, \ref{fig:corr_ab}, and \ref{fig:corr_fi}, we present the pairwise pearson correlation over instances for a few different values for each model's outputs on the full gold split.

Most of the localization metrics here seem to be somewhat correlated, although not as much as one might expect.
IOU seems to be generally more correlated with AUROC than with Average Precision.

Of particular note is the correlation between Attention Entropy and the global and local similarities: Attention Entropy is usually slightly positively correlated with Global Similarity and slightly negatively correlated with Local Similarity.
Though it is still unclear why this is, it may have to do with a model's ability to localize seeing as this is more pronounced in models that perform better localization.

Finally, it is interesting that \textbf{+Abnormal} model has a somewhat negative correlation between Attention Entropy and all of the localization metrics, potentially indicating a connection between examples of abnormalities and Attention Entropy, but more work should be done to probe this further.

\subsection{Precision and IOU at different Thresholds}

Finally, we present Precision (Table \ref{tab:precision}) and IOU (Table \ref{tab:iou}) at different thresholds to get a better sense for the differences in the attention between each model.
(Some IOU scores for GLoRIA are repeated here to allow for an easier comparison.)
It is also clear that the Masking Model performs the best when only taking the top 5 or 10 percent, but GLoRIA starts producing similar or better scores at less strict thresholds.
The precision scores above 70\% here for \textbf{+Masking}, which far exceed any other model's scores at any threshold, give the sense that this model is quite effective at localization, but the dropoff when looking at the subsets do indicate the need for future work in this area.

\begin{table*}[]
    \centering
    \resizebox{\textwidth}{!}{
    \begin{tabular}{l c c c c c}
\hline
Model & Synth & \textbf{All} & \textbf{Abnormal} & \textbf{One Lung} & \textbf{MDRB} \\
\hline
\multirow{2}{*}{UNITER*} & \xmark & 63.08/66.66/63.82 & 60.16/63.33/58.51 & 47.73/47.62/45.97 & 50.83/52.21/46.34 \\
 & \greencheck & 63.18/66.59/63.86 & 61.69/63.96/58.69 & 47.74/47.98/45.96 & 49.68/50.12/46.04 \\
\hline
\multirow{2}{*}{GLoRIA} & \xmark & 58.56/59.20/54.98 & 53.63/54.60/51.57 & 42.70/43.57/39.89 & 41.00/41.48/37.90 \\
 & \greencheck & 58.70/58.82/55.23 & 57.46/56.77/51.09 & 50.53/47.57/39.18 & 42.80/42.37/38.51 \\
\hline
\multirow{2}{*}{GLoRIA Retrained} & \xmark & 34.08/37.81/40.04 & 32.82/33.56/35.18 & 25.63/26.73/27.86 & 26.05/26.35/27.81 \\
 & \greencheck & 34.12/37.08/39.61 & 29.32/31.86/34.81 & 22.00/25.58/27.95 & 25.76/26.21/27.64 \\
\hline
\multirow{2}{*}{+Word Masking} & \xmark & 20.69/36.06/45.14 & 26.36/34.84/40.59 & 19.87/27.67/31.53 & 16.73/26.55/31.99 \\
 & \greencheck & 18.38/34.34/43.72 & 22.91/33.24/38.38 & 15.16/25.96/30.36 & 13.45/24.64/30.38 \\
\hline
\multirow{2}{*}{+Clinical Masking} & \xmark & 27.79/35.72/40.07 & 30.70/33.16/35.67 & 21.71/26.05/28.07 & 21.80/26.08/27.88 \\
 & \greencheck & 24.41/35.07/40.37 & 24.83/31.75/35.55 & 17.27/24.87/27.96 & 18.47/24.89/27.96 \\
\hline
\multirow{2}{*}{+``No Attn'' Token} & \xmark & 37.93/38.97/40.19 & 40.48/35.93/35.75 & 35.38/28.68/28.34 & 28.24/27.59/28.12 \\
 & \greencheck & 36.37/37.67/40.17 & 39.44/34.43/35.94 & 36.70/28.19/28.79 & 28.09/26.50/28.29 \\
\hline
\multirow{2}{*}{+Abnormal} & \xmark & 42.95/33.30/39.20 & 48.32/37.29/36.36 & 40.43/30.22/28.08 & 34.69/25.58/27.47 \\
 & \greencheck & 35.11/26.45/37.95 & 33.62/23.35/36.04 & 25.90/16.99/27.77 & 26.31/19.12/27.06 \\
\hline
\multirow{2}{*}{+30-shot Finetuned} & \xmark & 73.15/69.35/39.67 & 67.45/64.35/37.87 & 52.55/49.84/30.49 & 53.46/49.25/28.10 \\
 & \greencheck & 70.95/70.05/46.92 & 62.81/62.99/51.54 & 52.04/50.24/39.39 & 50.95/49.66/34.66 \\
\hline
\multirow{2}{*}{+Rand Sents} & \xmark & 14.54/14.98/23.22 & 15.66/15.37/22.26 & 11.66/11.61/17.05 & 9.31/10.18/16.61 \\
 & \greencheck & 8.68/8.94/20.00 & 13.62/12.92/21.15 & 4.78/4.32/12.75 & 4.67/5.68/14.81 \\
\hline
    \end{tabular}
    }
    \caption{\textbf{Precision} at 5/10/30\%}
    \label{tab:precision}
\end{table*}

\begin{table*}[]
    \centering
    \resizebox{\textwidth}{!}{
    \begin{tabular}{l c c c c c}
\hline
Model & Synth & \textbf{All} & \textbf{Abnormal} & \textbf{One Lung} & \textbf{MDRB} \\
\hline
\multirow{2}{*}{UNITER*} & \xmark & 2.57/7.17/33.61 & 2.56/6.95/34.88 & 2.83/7.81/30.78 & 3.14/9.21/30.33 \\
 & \greencheck & 2.71/7.54/34.13 & 2.76/8.36/35.53 & 2.88/8.03/31.85 & 3.73/9.77/30.50 \\
\hline
\multirow{2}{*}{GLoRIA} & \xmark & 3.79/6.69/20.10 & 4.10/7.25/19.05 & 4.43/8.05/20.54 & 3.56/6.37/16.92 \\
 & \greencheck & 4.89/8.96/23.62 & 7.20/13.25/29.30 & 7.55/12.82/27.69 & 4.83/8.24/19.84 \\
\hline
\multirow{2}{*}{GLoRIA Retrained} & \xmark & 2.51/3.80/4.21 & 3.10/3.86/4.08 & 2.29/2.87/3.14 & 3.27/4.68/4.82 \\
 & \greencheck & 2.75/4.21/4.74 & 3.43/3.89/4.21 & 2.33/2.77/3.21 & 3.25/4.82/5.01 \\
\hline
\multirow{2}{*}{+Word Masking} & \xmark & 1.79/2.60/3.48 & 2.77/3.69/4.34 & 2.43/3.19/3.90 & 2.14/2.83/3.40 \\
 & \greencheck & 1.55/2.26/3.06 & 2.50/3.11/3.48 & 1.81/2.31/2.88 & 1.54/2.24/2.65 \\
\hline
\multirow{2}{*}{+Clinical Masking} & \xmark & 1.63/2.18/2.54 & 2.66/3.12/3.26 & 1.36/1.58/1.73 & 1.89/2.54/2.67 \\
 & \greencheck & 1.65/2.17/2.72 & 2.19/2.45/2.70 & 1.14/1.45/1.78 & 2.18/2.59/2.93 \\
\hline
\multirow{2}{*}{+``No Attn'' Token} & \xmark & 3.13/4.17/4.32 & 5.22/6.54/6.59 & 6.01/7.43/7.64 & 3.82/5.19/5.29 \\
 & \greencheck & 3.09/4.06/4.29 & 5.04/5.97/6.22 & 6.07/6.96/7.26 & 3.24/4.80/4.93 \\
\hline
\multirow{2}{*}{+Abnormal} & \xmark & 5.45/8.08/9.08 & 6.60/10.48/10.49 & 6.28/9.45/9.76 & 6.12/7.94/8.05 \\
 & \greencheck & 4.71/6.20/7.24 & 5.51/6.72/6.91 & 4.46/5.09/6.00 & 4.90/5.97/6.09 \\
\hline
\multirow{2}{*}{+30-shot Finetuned} & \xmark & 9.53/18.24/30.05 & 9.91/18.83/31.24 & 9.30/17.50/27.86 & 9.01/16.09/25.24 \\
 & \greencheck & 9.44/18.01/34.38 & 9.23/17.54/38.35 & 9.37/16.98/32.41 & 8.59/15.57/27.89 \\
\hline
\multirow{2}{*}{+Rand Sents} & \xmark & 0.35/0.76/5.51 & 0.36/0.62/4.85 & 0.43/0.76/4.68 & 0.16/0.59/4.46 \\
 & \greencheck & 0.45/0.94/7.35 & 0.47/0.75/5.90 & 0.66/1.11/7.18 & 0.22/0.70/6.22 \\
\hline
    \end{tabular}
    }
    \caption{\textbf{IOU} at 5/10/30\%}
    \label{tab:iou}
\end{table*}

\end{document}